\documentclass[11pt,letterpaper]{yalearxiv}

\graphicspath{{figs/}}
\usepackage{xr}
\usepackage{xcite}
\usepackage{subfiles}
\usepackage{booktabs}
\usepackage{xurl}
\usepackage[caption=false]{subfig}
\usepackage[T1]{fontenc}
\usepackage{lmodern}
\usepackage{textcomp}
\usepackage{lineno}
\usepackage{graphicx}
\usepackage{caption}
\usepackage{bbm}
\usepackage{rotating}
\usepackage{tikz}
\usetikzlibrary{positioning,arrows.meta}
\usetikzlibrary{calc}
\usepackage{enumitem}
\newcommand\blfootnote[1]{%
  \begingroup
  \renewcommand\thefootnote{}%
  \footnote{#1}%
  \addtocounter{footnote}{-1}%
  \endgroup
}
\usetikzlibrary{positioning,arrows.meta}

\makeatletter

\newcommand*{\addFileDependency}[1]{
\typeout{(#1)}
\@addtofilelist{#1}
\IfFileExists{#1}{}{\typeout{No file #1.}}
}\makeatother

\newcommand*{\myexternaldocument}[1]{%
\externaldocument{#1}%
\addFileDependency{#1.tex}%
\addFileDependency{#1.aux}%
}
\myexternaldocument{./SI}

\begin{document}

\title{EpiFlow: A framework for improving the utility of wastewater signals for disease forecasting}

\author{
Aniruddha Adiga$^{1,*}$, Jingyuan Chou$^{1,2}$, Gursharn Kaur$^{1}$, Andrew Warren$^{1}$, Srinivasan Venkatramanan$^{1}$, Baltazar Espinoza$^{1}$, Bryan Lewis$^{1}$, Justin Crow$^{3}$, Alexandra Lorentz$^{4}$, Rekha Singh$^{5,6}$ and Madhav Marathe$^{1,2,*}$}

\blfootnote{$^{1}$Biocomplexity Institute, University of Virginia\\
$^{2}$Department of Computer Science, University of Virginia\\
$^{3}$Virginia Department of Health\\
$^{4}$Department of General Services, Division of Consolidated Laboratory Services\\
$^{5}$Biodefense Fellow, US Department of Defense, 
Pentagon\\
$^{6}$Department  of Civil and Environmental Engineering, Old Dominion University\\
$^{*}$Corresponding authors: aniruddha@virginia.edu, marathe@virginia.edu
}



\begin{abstract}
    Wastewater-based surveillance is an effective tool for disease monitoring, offering early warnings of outbreaks. While wastewater viral loads (WVL) correlate with disease burden, their utility in improving real-time forecasting remains under investigation. During early epidemic phases, many indicators effectively monitor spread, but their reliability may decline due to fatigue and low prevalence. Hospital burden can vary significantly even during low-prevalence periods—making accurate forecasting of burden indicators essential to minimizing impact.

In this paper, we present principled approaches for processing wastewater data, understanding its relationship with burden indicators, and generating real-time forecasts. We assess the predictability of WVL using entropy measures. We analyze the relationship between WVL and burden indicators using causality tests that capture temporal dynamics and WVL's leading-indicator behavior. We incorporate these insights into a time-varying forecasting model that accounts for the evolving relationship between signals. We also test the effects of delays in WVL reporting through simulations. We test utility of our methods by forecasting COVID-19 hospital admissions across Virginia and its health regions across different prevalence periods. Incorporating WVL improves forecast accuracy over baseline models, especially during critical phases, we observe a 20 percentage points improvement in forecast coverage. Our results show that WVL signals are important for infectious disease forecasting even under conditions of low prevalence or reporting delays.
\end{abstract}
%
\maketitle

\section{Introduction}
Wastewater-based surveillance has become an important complement to clinical surveillance for monitoring infectious diseases because it can capture infections across symptomatic and asymptomatic populations while being less affected by healthcare-seeking behavior~\cite{nwss,national2023phase1}. However, wastewater viral load (WVL) signals are subject to noise and uncertainty due to variable catchment populations, shedding dynamics, flow rates, dilution, wastewater properties, environmental conditions, and sampling strategies~\cite{bertels2023time,greenwald2021tools,mohring2024estimating}. These uncertainties complicate the direct use of WVL for real-time forecasting of burden indicators such as hospital admissions.

A useful test of wastewater utility is whether it improves forecasts of disease burden beyond models that use burden indicators alone. This question is especially important as COVID-19 transitions toward lower-prevalence and post-public-health-emergency periods, when clinical testing behavior changed substantially and WVL can exhibit reduced signal-to-noise ratio~\cite{burki2023ends,emergency_term}. Although several studies have shown that SARS-CoV-2 wastewater concentrations can lead reported cases and hospitalizations~\cite{feng2021evaluation,phan2023simple,weidhaas2021correlation,jiang2022artificial,duvallet2022nationwide,keshaviah2022separating,nattino2022association,zhan2022relationships,kaplan2021aligning,galani2022sars,peccia2020measurement}, the Phase 2 NASEM report emphasizes that the use of wastewater data for operational real-time forecasting remains at an early stage and requires improved data quality and integration methods~\cite{national2024increasing}.

Here, we present \emph{EpiFlow}, an integrated framework for improving the real-time forecasting utility of wastewater signals. EpiFlow combines data preprocessing, signal-quality analysis, dynamic relationship assessment, and probabilistic forecasting. The framework converts sewershed-level WVL into viral activity levels (VAL), evaluates intrinsic signal predictability using permutation entropy, analyzes time-varying dependencies between wastewater and burden indicators using rolling-window Granger causality (RWGC), and generates probabilistic forecasts using a rolling vector autoregressive (VAR) model. The framework also evaluates the effect of wastewater reporting delays, which commonly arise from sampling and testing workflows.

\paragraph{Summary of results.} We demonstrate EpiFlow framework using SARS-CoV-2 wastewater data and COVID-19 hospital admissions across five Virginia health regions and the state level from October 2021 to December 2023. Forecasts are evaluated against ARIMA and ARIMAX baselines using weighted interval score (WIS) and forecast coverage, which are standard metrics in epidemic forecasting~\cite{bracher2021evaluating,mathis2024evaluation}. Our results show that wastewater-derived indicators can improve probabilistic forecasts, particularly when denoised and modeled through time-varying VAR dynamics. The improvement is most pronounced during surge periods and remains useful even when wastewater reporting is delayed.

\section{Related Work}
Wastewater-based epidemiology has been used to track SARS-CoV-2 and other pathogens, and multiple studies have shown associations between wastewater concentrations and clinical indicators such as cases, hospital admissions, and deaths~\cite{feng2021evaluation,phan2023simple,weidhaas2021correlation,jiang2022artificial,duvallet2022nationwide,keshaviah2022separating,nattino2022association,zhan2022relationships,kaplan2021aligning,galani2022sars,peccia2020measurement}. Statistical and data-driven models used in this area include ARIMA, VAR, generalized additive models, negative binomial models, Bayesian filters, state-space models, semi-mechanistic models, random forests, nearest-neighbor methods, and neural networks~\cite{karthikeyan2021high,zhao2022five,cao2021forecasting,galani2022sars,chen2024wastewater}. However, several of these studies are restricted to short time frames, conducted during the early stages of the pandemic, and limited to a single or few districts with a relatively high sampling rate of WVL in sewersheds (daily, thrice weekly, etc.).

A recurring challenge is that WVL signals are noisy and heterogeneous, which can reduce interpretability and forecasting performance. Prior work has used Bayesian smoothing, signal processing, and state-space methods to mitigate uncertainty and recover stable temporal trends~\cite{dai2022statistical,greenwald2021tools,mohring2024estimating}. Recent wastewater studies further emphasize dynamic and state-space modeling for handling missing observations, citywide trend assessment, and time-varying relationships between wastewater concentrations and infections~\cite{CDCWastewaterForecast2024,ensor2025nonlinear,sun2025uncovering}. In contrast to studies focused primarily on point forecasts or high-prevalence periods, EpiFlow evaluates whether wastewater signals improve real-time probabilistic forecasts across heterogeneous regions and across both PHE and post-PHE periods.

The references cited in this paper is not an exhaustive list. We refer the reader to SI~\ref{sec:related_works_SI} and Chen-Chen et al.~\cite{chen2024wastewater} for a detailed review of the various methods and models employed in wastewater-based epidemiology. 

\section{EpiFlow Framework and Methods}
\label{sec:methods_condensed}
\begin{figure}[t]
    \centering
\includegraphics[width=.92\linewidth]{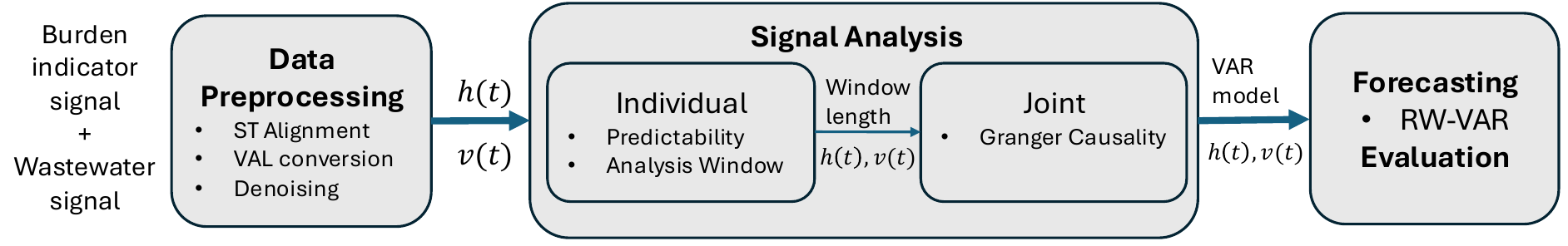}
    \caption{Overview of the EpiFlow framework. The framework consists of data preprocessing, signal analysis, and forecasting modules.}
    \label{fig:pipeline}
\end{figure}

EpiFlow is organized into three modules to address several challenges experienced when working with wastewater data and burden indicators: data preprocessing, signal analysis, and forecasting (Figure~\ref{fig:pipeline}). The preprocessing module aligns wastewater and hospitalization signals in space and denoises the wastewater signal; the signal analysis module quantifies wastewater predictability and dynamic relationships with burden indicators; and the forecasting module generates probabilistic forecasts.
\subsection{Data Preprocessing and Signal Construction}
\subsubsection{Data Preparation}
Wastewater and hospital admission data are first temporally aligned to weekly resolution. Daily hospital admissions are aggregated to weekly counts and assigned to the beginning of the corresponding epidemiological week. For wastewater viral load (WVL) measurements, once-weekly samples are assigned to the corresponding week; when two samples are available in a week, we compute the average of the log-transformed viral concentrations and assign the resulting value to that week. Missing values are imputed using forward filling from the most recent available observation. Data sources containing more than four consecutive missing values are excluded from the analysis. Spatial alignment is performed at the Virginia health-region level. Hospital admissions are summed across facilities serving each region, while sewershed-level WVL signals are transformed into viral activity levels (VAL) following NWSS-specified normalization~\cite{nwss} (See SI~\ref{sec:VAL-SI}). The regional VAL is computed as the median VAL across sewersheds serving a region. This produces weekly paired time series of hospital admissions, denoted by $h(t)$, and wastewater-derived viral activity, denoted by $v(t)$.
\subsubsection{Denoising and Savitzky--Golay Filter Selection}
Because WVL and VAL signals can contain high-frequency fluctuations due to sampling variability, catchment heterogeneity, flow effects, and reporting irregularities, we additionally analyze a denoised VAL signal, denoted VAL-dn. Denoising is performed using a Savitzky--Golay (SG) filter, which smooths a time series by fitting a low-degree polynomial locally within a moving window and evaluating the fitted polynomial at the center point~\cite{schafer2011savitzky,krishnan2012selection}. The SG filter is controlled by two parameters: the polynomial order $m$ and the filter window length $L$. 


To select a practical denoising configuration, we swept over candidate polynomial orders $m\in\{1,2,3\}$ and candidate filter lengths $L\in\{1,3,\ldots,11\}$ (odd values only). For each candidate pair $(m,L)$ and each regional VAL time series, the filtered signal was computed and compared against the original signal using mean squared error (MSE). The objective was not to remove all high-frequency variation, but to identify a filter that reduces short-scale noise while preserving the timing and amplitude of epidemiologically meaningful changes. Denoising is performed each time a new observation is obtained. Since SG filters are zero-phase filters, polynomial interpolation was used near the signal boundaries instead of explicit padding, thereby reducing edge artifacts and preserving local temporal structure of the most recent observations. See SI~\ref{app:s-g_filters} for more details on SG filters.
\subsection{Signal Analysis}
\subsubsection{Predictability Analysis using Permutation entropy}
Permutation entropy (PE) is used to quantify temporal structure and intrinsic predictability in the wastewater and hospitalization signals~\cite{bandt2002permutation}. PE is a model-free ordinal measure of uncertainty that summarizes the frequency of rank-order patterns within a time series. For a rolling window\footnote{A window refers to a fixed-length sequence of observed values. A forecast horizon denotes the length of time into the future for which predictions are generated.} of observations, embedding dimension $d$, and delay $\tau$, the time series is mapped to ordinal patterns of length $d$; the Shannon entropy of the empirical ordinal-pattern distribution is then normalized by $\log(d!)$. We define predictability as $1-\mathrm{PE}$, so that higher values indicate stronger ordinal structure and lower uncertainty. In~\cite{scarpino2019predictability}, PE was employed to understand the predictability of multiple infectious disease time series. Infectious disease time series are affected by observation errors which introduces uncertainties and reduces the redundant information present in the signal. Using a rolling-window-based analysis, it was observed that as the window length increases, the predictability of the signal drops. An important observation which indicates that due to the uncertainties, using long periods of the time series (more data points) does not necessarily help forecast the future values of the signal. It was shown that the predictability drops significantly beyond 10$-$16 weeks.

\paragraph{Observation Window.} In our analysis, PE is computed within rolling windows using embedding dimensions $d=3$ and $d=4$ and unit delay $\tau=1$. Higher embedding dimensions are avoided because short rolling windows contain too few observations to reliably estimate all $d!$ ordinal patterns. To determine an appropriate rolling-window length for downstream analysis, we compare mean predictability across candidate window lengths ranging from short local windows to the full available time series. For each candidate length, predictability is computed across all rolling windows and locations. We then compare mean predictability across window lengths using Tukey's Honest Significant Difference (HSD) test at $\alpha=0.05$ (see SI~\ref{sec:predictability-SI}) and select the smallest window length beyond which increasing the window no longer produces statistically significant changes in mean predictability. This procedure identifies a window size that balances local adaptivity with sufficient sample size for low entropy and model estimation.
\paragraph{Permutation-based significance testing for permutation entropy.} To evaluate whether the observed PE over relatively short window lengths reflects non-random temporal structure, we use a random-shuffle permutation test. For each rolling window, the observed normalized entropy $H_{\mathrm{obs}}$ is compared with an empirical null distribution obtained from $B=500$ randomly permuted (shuffled) surrogate series. Permuting preserves the marginal distribution of the time series while destroying temporal dependence~\cite{Theiler1992,Good2005}. Since structured signals typically have lower entropy than randomly ordered sequences, we use a one-sided test with empirical $p$-value
    $p = \frac{1}{B}\sum_{b=1}^{B}\mathbb{I}\left(H^{(b)}\leq H_{\mathrm{obs}}\right),$
where $H^{(b)}$ is the PE of the $b$th shuffled surrogate. If $p<0.05$, then PE value for the particular window is considered statistically significant.

\subsection{Rolling-Window Granger Causality Analysis}
Prior to discussing the methods to capture the relationship between hospital admissions $h(t)$ and VAL $v(t)$, it is to be noted that wastewater viral load primarily reflect infection prevalence, whereas hospital admissions represent incidence. Consequently, the two signals may exhibit indirect but potentially useful predictive relationships for forecasting. We refer the reader to SI~\ref{sec:GC_all-SI} for a detailed description along with a conceptual causal graph.

Rolling-Window Granger causality (RWGC), with window length obtained from predictability analysis, is used to characterize time-varying directional dependencies between $h(t)$ and $v(t)$. The RWGC analysis is based on a bivariate vector  (VAR) model fitted independently within each rolling window. For a VAR($P$) model,
\begin{align}
\begin{bmatrix}
h(t)\\
v(t)
\end{bmatrix}
&=
\begin{bmatrix}
c_h\\
c_v
\end{bmatrix}
+
\sum_{p=1}^{P}
\begin{bmatrix}
a^{(p)}_{hh} & a^{(p)}_{hv}\\
a^{(p)}_{vh} & a^{(p)}_{vv}
\end{bmatrix}
\begin{bmatrix}
h(t-p)\\
v(t-p)
\end{bmatrix}
+
\mathbf{e}(t),
\label{eq:var_granger}
\end{align}
where $\mathbf{e}(t)$ is a zero-mean residual process. In the bivariate setting, the first row of the coefficient matrix governs the hospitalization equation and the second row governs the VAL equation. Thus, the coefficients $a^{(p)}_{hv}$ quantify the contribution of lagged VAL to current hospital admissions, while $a^{(p)}_{vh}$ quantify the contribution of lagged hospital admissions to current VAL.

For each rolling window, the VAR design matrix is constructed from the lagged observations $\mathbf{Y}(t-1),\ldots,\mathbf{Y}(t-P)$, where $\mathbf{Y}(t)=[h(t),v(t)]^{\top}$. The two response equations are estimated independently using ordinary least squares. Specifically, the coefficients are obtained by minimizing the residual sum of squares within the training window, yielding estimates of the lag coefficient matrices and the residual covariance matrix. Based on multiple observations~\cite{hill2023wastewater}, the lag order is fixed to $P=2$ in the analysis which reflects the short reporting and epidemiological delays expected between wastewater and hospitalization signals, while limiting over-parameterization in short rolling windows.

To assess whether $v$ Granger-causes $h$, the null hypothesis is
$
H_0^{v\rightarrow h}: a^{(p)}_{hv}=0,\quad \forall p=1,\ldots,P,$
which states that lagged VAL does not improve prediction of hospital admissions after conditioning on past hospital admissions. Similarly, the null hypothesis for no causal influence from $h$ to $v$ is
$
H_0^{h\rightarrow v}: a^{(p)}_{vh}=0,\quad \forall p=1,\ldots,P.
$
Classical GC significance is assessed using the chi-square statistic from the nested-model comparison between a restricted model, in which the cross-lagged coefficients are set to zero, and an unrestricted model containing all lagged terms.
\paragraph{Permutation-based significance testing for RWGC.} Since asymptotic chi-square tests may be unreliable in short and noisy rolling windows, we additionally use the surrogate-based permutation test. Under the null hypothesis of no directed dependence, the predictor VAL series is permuted to break temporal alignment between predictor and response while preserving marginal distribution of the predictor. For each surrogate realization $b=1,\ldots,B$, the VAR model is refitted and the GC statistic $T^{(b)}$ is recomputed. The empirical null distribution $\{T^{(1)},\ldots,T^{(B)}\}$ is then compared with the observed statistic $T_{\mathrm{obs}}$ using
$
    p = \frac{1}{B}\sum_{b=1}^{B}\mathbb{I}\left(T^{(b)}\geq T_{\mathrm{obs}}\right).$
A $p<0.05$ indicates that the observed directional relationship is unlikely under the surrogate null distribution. The rolling implementation records both the direction of significant GC and the lag at which the strongest statistic is observed, enabling identification of time-varying leading-indicator behavior.

\subsection{Forecasting Models}
The rolling-window forecasting module uses the most recent training window to fit a model and then generates forecasts for future horizons. All models are re-estimated independently at each forecast origin using only data that would have been available at that time. This rolling-window design allows model parameters to adapt to changing epidemic dynamics.

\paragraph{VAR forecasting.}
For the VAR model, the same bivariate structure used in the GC analysis is used for forecasting hospital admissions. After fitting the VAR($P$) model within the window, forecasts are generated recursively across future horizons. The fitted VAR provides a predictive mean vector and a forecast error covariance matrix at each horizon, obtained by propagating the estimated residual covariance through the VAR recursion. Assuming approximately Gaussian forecast errors, the predictive distribution is summarized by the forecast mean and variance for the hospitalization component. Predictive quantiles are then extracted at the 23 quantile levels specified by the COVID-19 Forecast Hubs and FluSight~\cite{COVID-Hub,mathis2024evaluation}.

We compare the VAR forecasts against autoregressive integrated moving average (ARIMA) and ARIMAX (ARIMA with eXogenous variables) baselines for hospital admissions $h(t)$ forecasting. The ARIMA model uses only the history of $h(t)$, while the ARIMAX model additionally incorporates wastewater-derived viral activity $v(t)$ as an exogenous predictor using the current and lagged VAL values. Similar to the VAR framework, both models are trained using a rolling-window approach, where the training window is advanced by one step as new observations become available and the model is refit to the updated window. Model orders are selected automatically using the \texttt{AutoARIMA} framework from the \texttt{pmdarima} package~\cite{smith2017pmdarima,Hyndman2008}, which determines the optimal ARIMA configuration by minimizing the Akaike Information Criterion (AIC) over candidate model orders using a stepwise search procedure. For all models, multi-step probabilistic forecasts are generated recursively, and the resulting predictive distributions are converted into the same 23 quantiles used across all forecasting models. For ARIMAX forecasting, future wastewater values are first forecast and then supplied as exogenous inputs to the model. The ARIMAX model using forecast wastewater inputs is denoted ARIMAX-Fct, while the version using denoised VAL signals is denoted ARIMAX-Fct-dn.

\paragraph{Forecast quantiles and evaluation.}
For each model, forecast distributions are converted into 23 quantile levels used by the Forecast Hubs~\cite{mathis2024evaluation,COVID-Hub}.
If $\hat{\mu}_{t+h}$ and $\hat{\sigma}_{t+h}$ denote the forecast mean and standard deviation for hospitalization at horizon $h$, the quantile at probability level $\tau$ is computed as $\hat{q}_{\tau,t+h}=\hat{\mu}_{t+h}+z_{\tau}\hat{\sigma}_{t+h}$, where $z_{\tau}$ is the corresponding standard normal quantile. Forecast performance is evaluated using weighted interval score (WIS), scaled relative WIS with ARIMA as the baseline, and empirical coverage (see SI~\ref{app:forecast_eval} for definitions).

\section{Results: A Case Study on COVID-19 Hospital Admissions in Virginia}
\label{sec:case_study_condensed}
\begin{figure}[h!]
    \centering
    \includegraphics[width=0.5\linewidth]{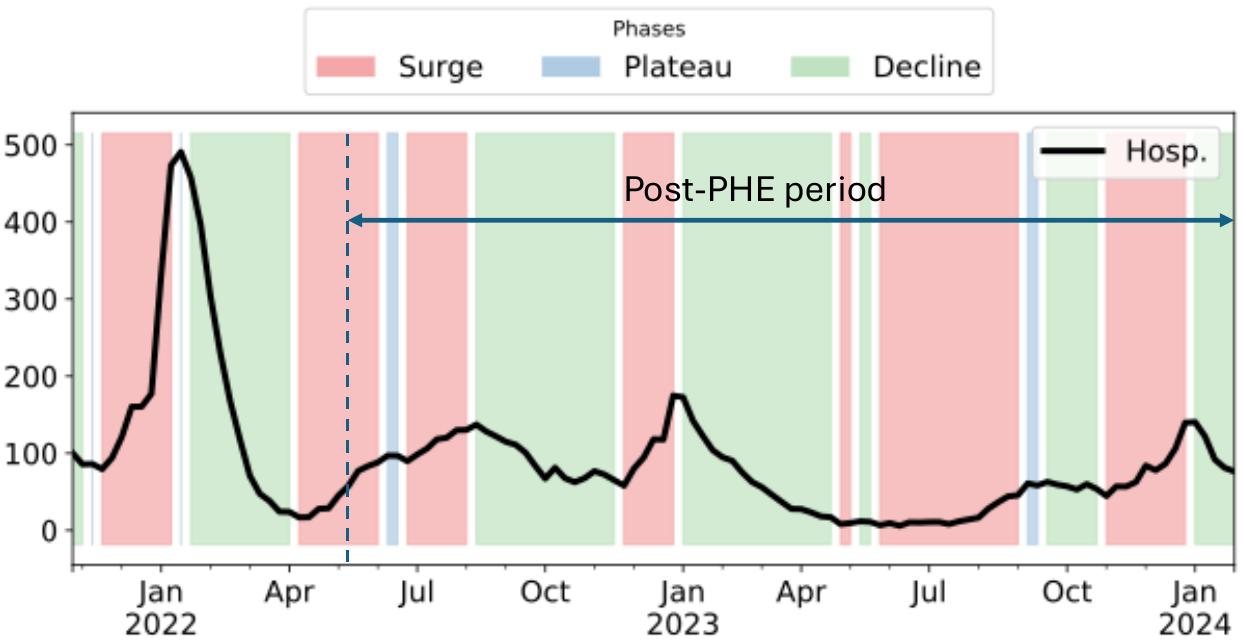}
    \caption{Classification of VA Hosp. time series weeks into phases. The weeks are shaded based on the phase: surge=soft red, plateau=soft blue, and decline=lime green.}
    \label{fig:VA-phase-classificatio}
\end{figure}

We evaluate EpiFlow using SARS-CoV-2 wastewater surveillance data from the Virginia Department of Health and COVID-19 hospital admissions from October 2021 to December 2023~\cite{VDHdash,hosp-hgov}. Wastewater data were collected from 36 treatment plants, with 13 sites sampled twice weekly and 23 sampled once weekly. Hospital admissions were aggregated to five Virginia health regions (Central, Eastern, Northern, Northwest, and Southwest) and to the state level (VA). WVL measurements were transformed to VAL, aggregated regionally, and denoised (VAL-dn). To evaluate performance during epidemiologically meaningful periods, each week was classified into surge, plateau, or decline phases using a piecewise-linear approximation (See SI~\ref{sec:phase-class}) of the hospital admission time series (Figure~\ref{fig:VA-phase-classificatio}). Phase-specific evaluation was used to assess model behavior during critical periods.

\subsection{Predictability of Wastewater Signals}
\label{sec:predictability_results}
\begin{figure}[h!]
\centering
    \subfloat[]{
        \includegraphics[width=0.45\textwidth]{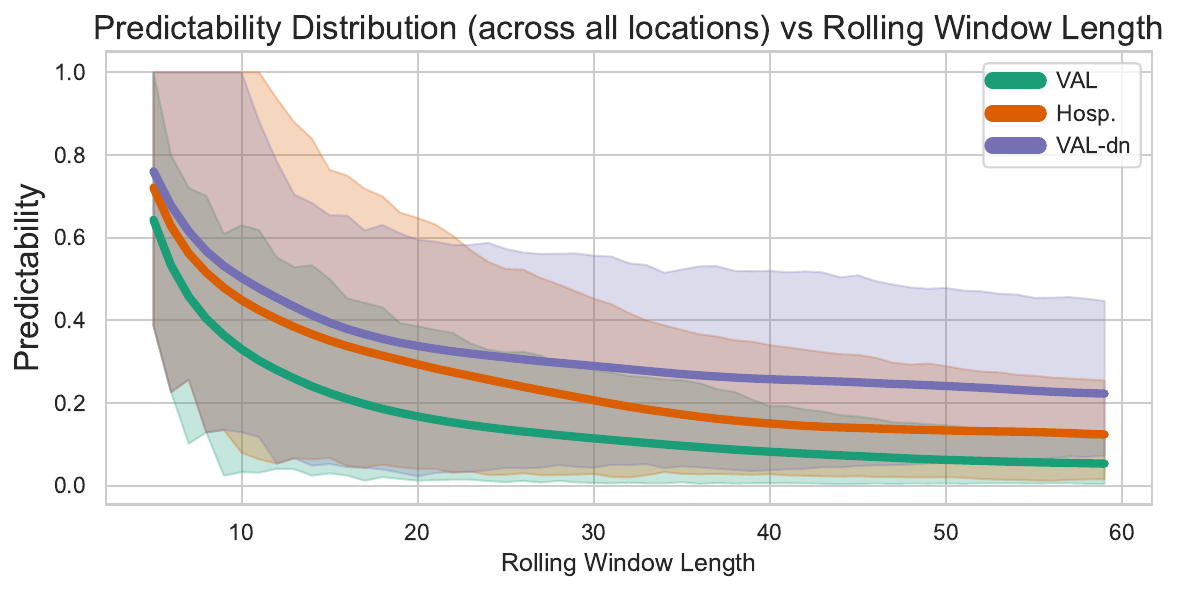}
        \label{fig:pred_distn_al_reg}}
    \subfloat[]{
        \includegraphics[width=.45\textwidth]{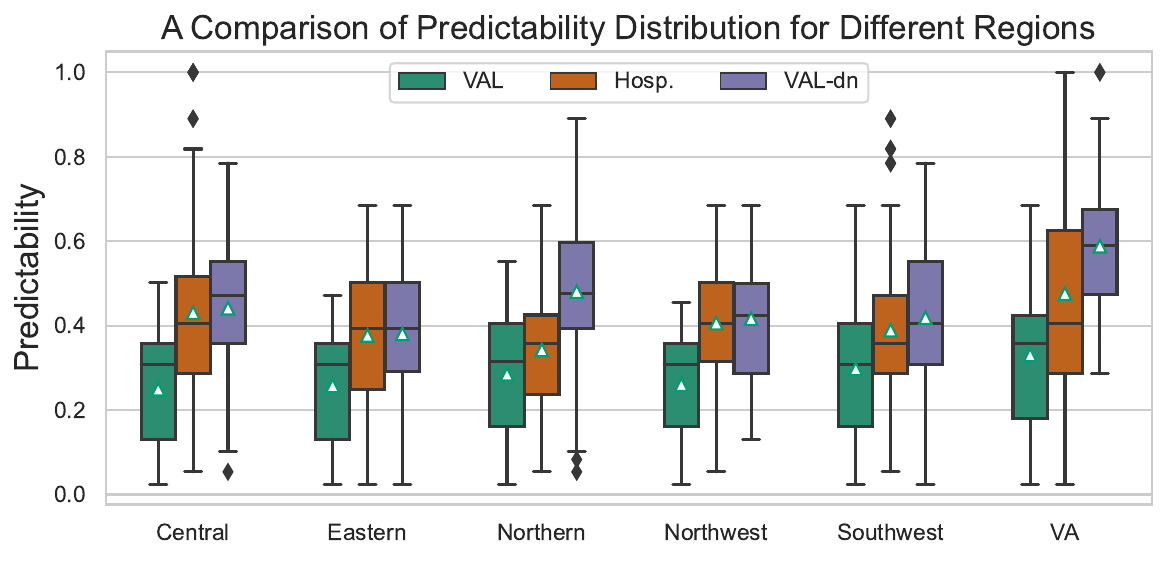}
        \label{fig:perm_entropy_dist}}
    \caption{Predictability of VAL, COVID-19 hospitalization, and denoised VAL. PE order restricted to 3 and 4 (significance of PE values determined using the permutation test). (a) Analyzing the distribution of predictability for different rolling window lengths (lines indicate the mean and the bands indicate the interquartile range of the distribution). (b) Distribution of predictability for a window length of 12 across different regions. Overall, we observe that for all the three time series, the predictability drops as window length increases. The VAL time series has considerably lower predictability compared to the COVID-19 hospitalization time series. However, by denoising we observe significant improvement in predictability of the VAL time series (significance of the improvement established using two-sample KS test - see SI~\ref{sec:KS-test_SI}).}
    \label{fig:perm_entropy}
\end{figure}

Figure~\ref{fig:perm_entropy} summarizes the rolling-window predictability analysis. Predictability decreases as window length increases, indicating that longer histories do not necessarily improve usable temporal structure of noisy wastewater signals. VAL is consistently less predictable than hospital admissions, reflecting the high uncertainty in wastewater observations. However, denoising improves VAL predictability across regions, supporting the use of VAL-dn in forecasting models. Overall, we observed the window lengths varied between $10-16$ across locations and weeks, similar to the observations made in~\cite{scarpino2019predictability}.

\subsection{Time-Varying Granger Relationships}
\begin{figure}[h!]
\centering
\includegraphics[width=.8\textwidth]{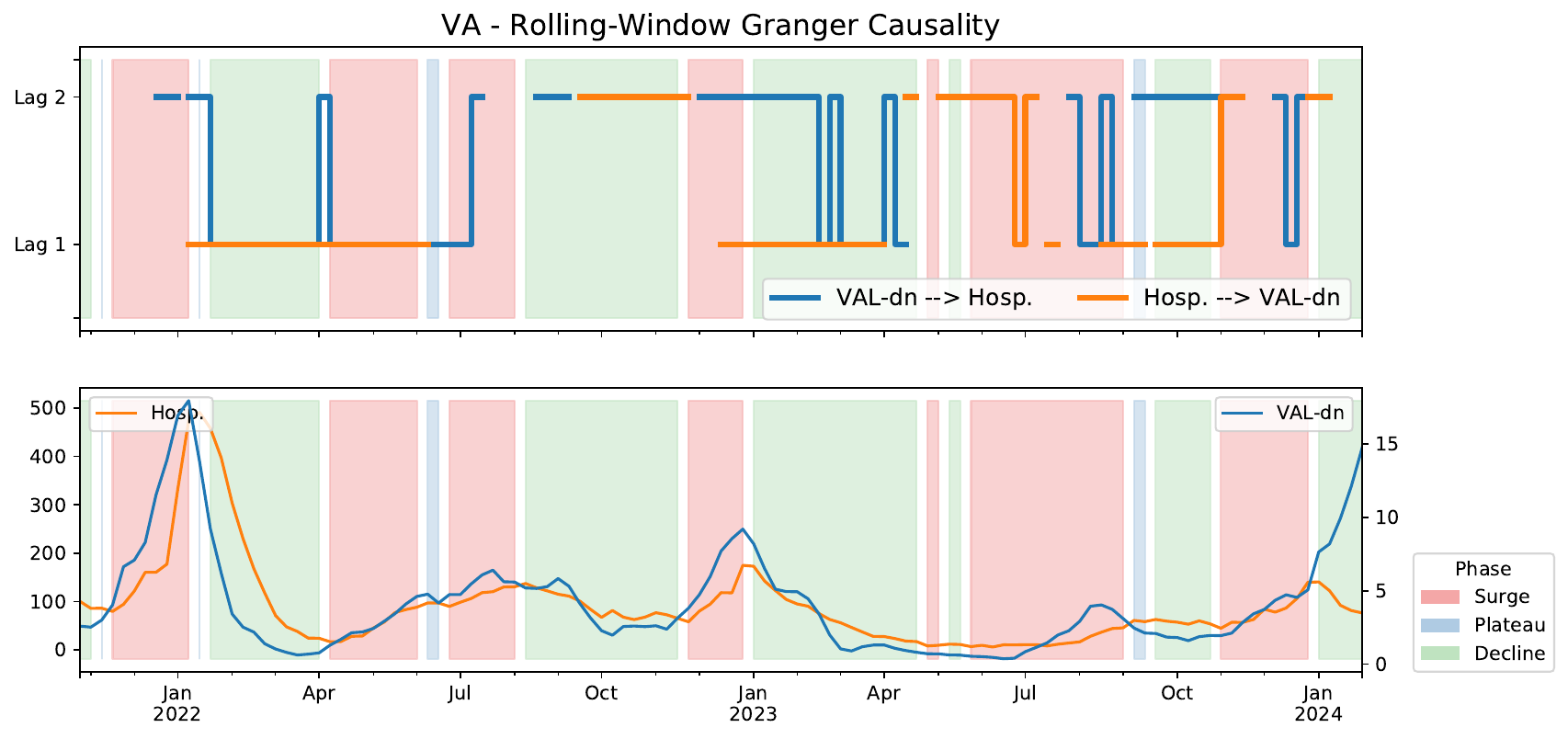}
    \caption{A rolling-window Granger causality test (using permutation-based significance test) to understand the dependence between hospitalizations and VAL-dn across time. In the top plot, the line color indicates the variable Granger causing the other variable and the value indicates the corresponding lag for that week. The bottom plot shows the VAL and Hosp. time series.  GC test indicates that VAL-dn Granger causes Hosp. for certain weeks (blue line in the top plot), while for other weeks, Hosp. Granger causes VAL-dn (orange line in the top plot). Mostly, during a surge phase, VAL-dn leads hospitalizations by 1--2 weeks.}
    \label{fig:VA-granger_test}
\end{figure}

Rolling-window GC analysis reveals that the relationship between VAL-dn and hospital admissions changes over time (Figure~\ref{fig:VA-granger_test}). During surge periods, wastewater-derived viral activity often Granger-causes hospital admissions, consistent with leading-indicator behavior. In other periods, the direction can weaken, reverse, or become bidirectional, indicating that the wastewater--hospitalization relationship is not stationary over the study period.

\subsection{Forecasting Performance}
\begin{figure}[h!]
\centering
    \subfloat[]{
        \includegraphics[width=0.45\textwidth]{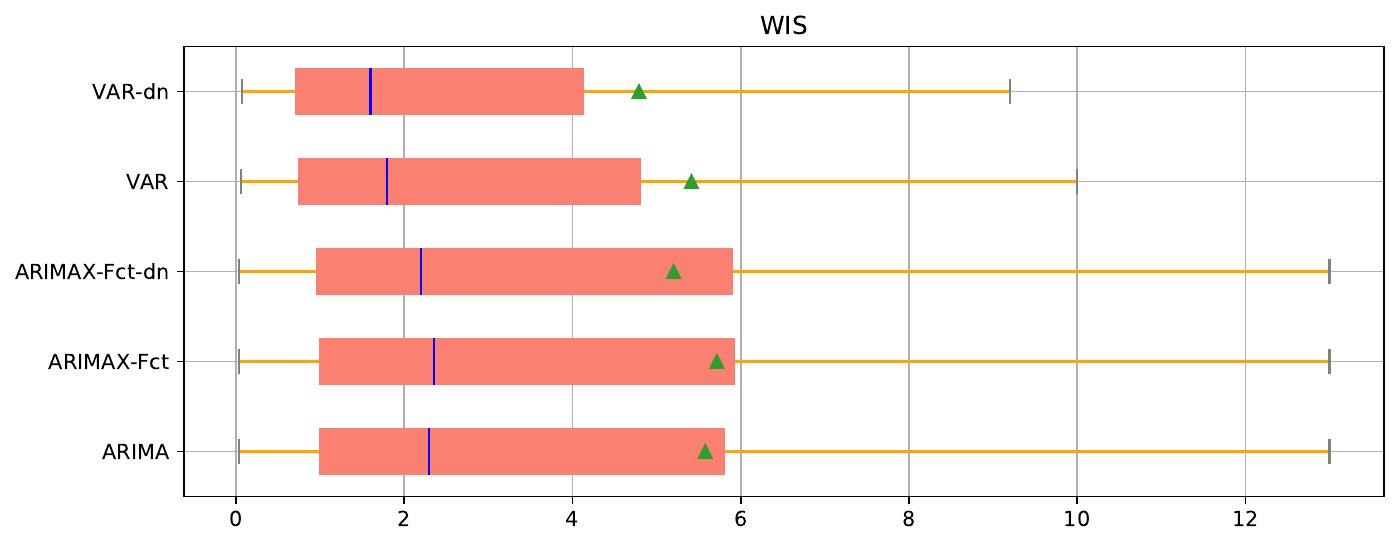}
        \label{fig:boxplot_wis}}
    \subfloat[]{
        \includegraphics[width=0.45\textwidth]{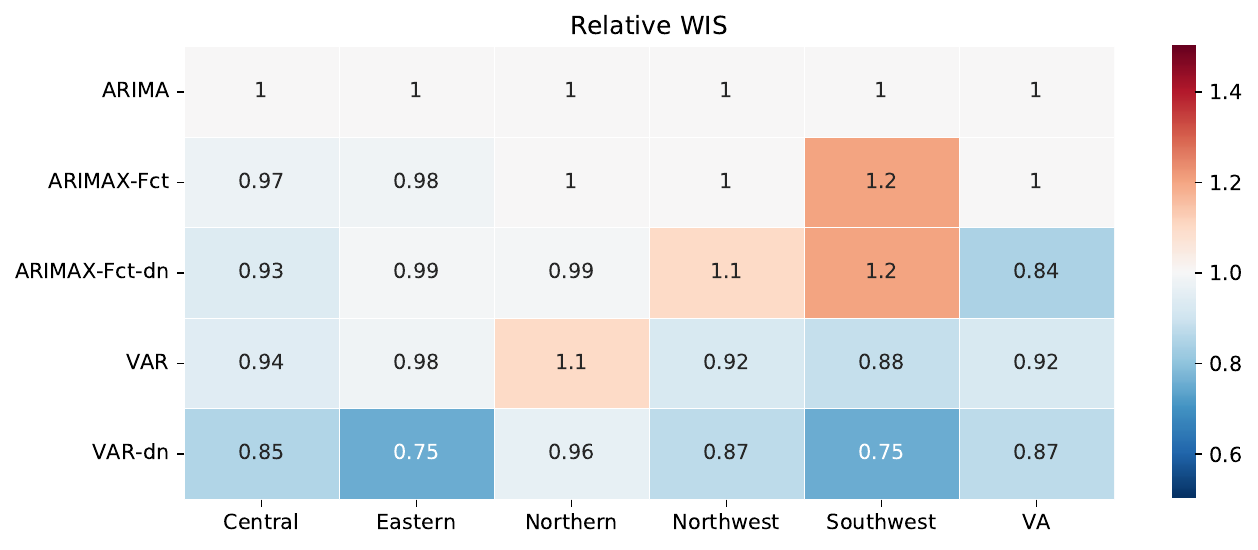}
        \label{fig:rel_wis_scores}}
        \\
   \subfloat[]{\includegraphics[width=.85\textwidth]{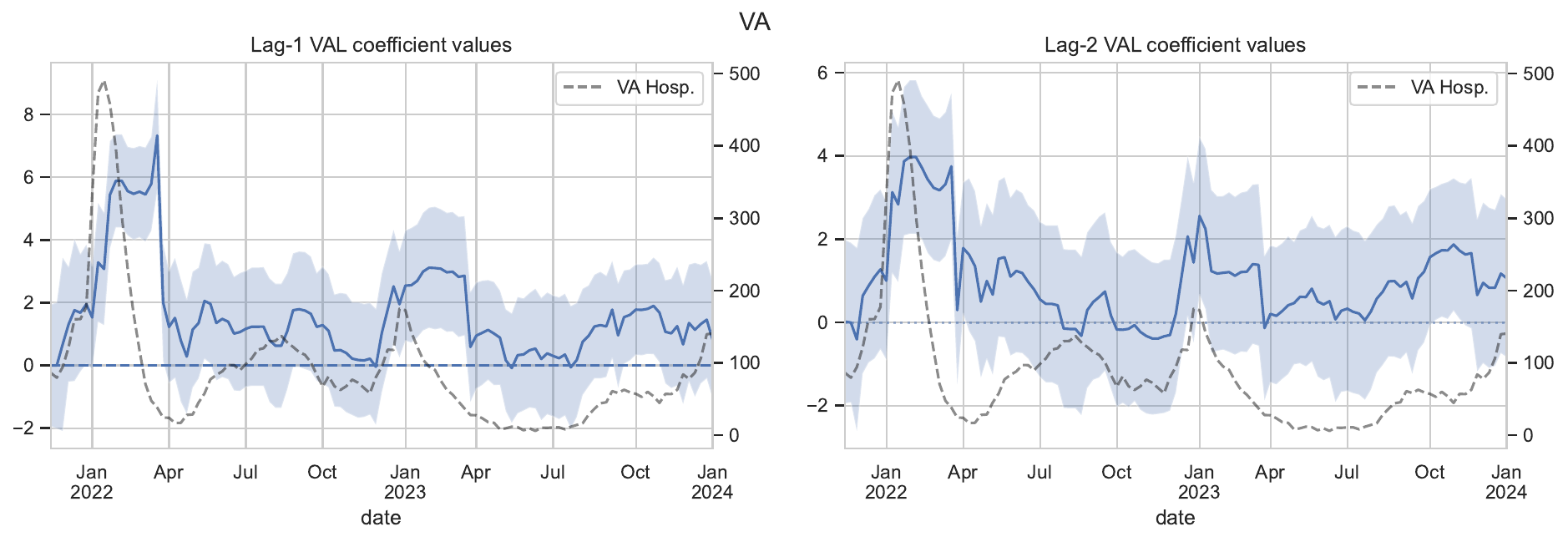}
        \label{fig:lags} }
        \\
    \subfloat[]{
    \includegraphics[width=0.75\textwidth]{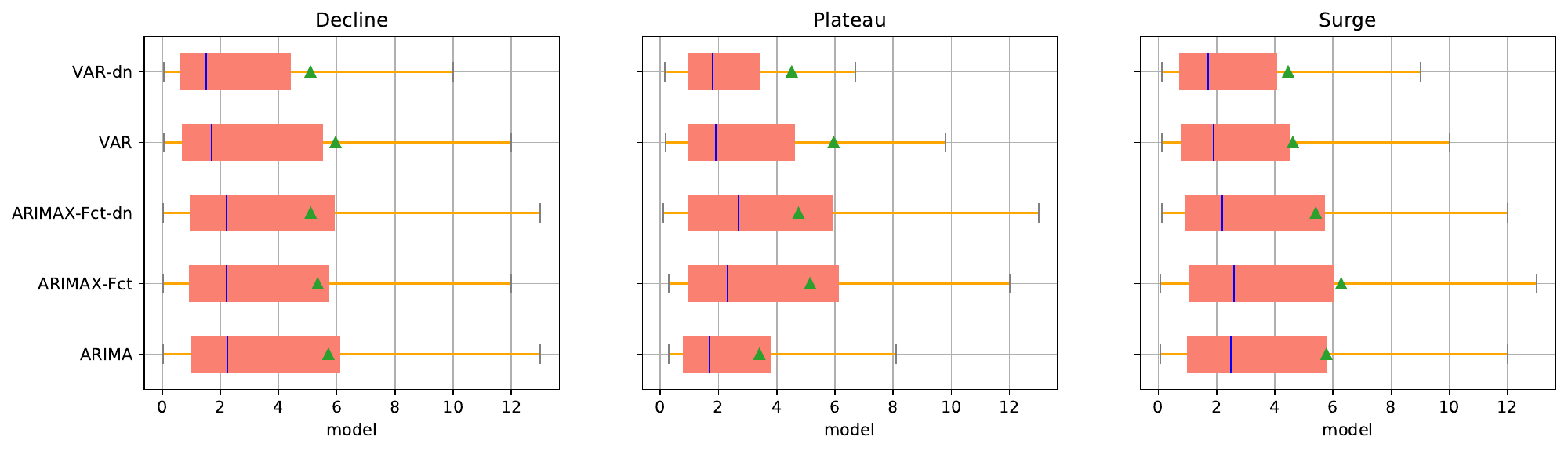}
            \label{fig:phase_wis}}
    \caption{Summary of the WIS scores for each method (post-PHE period). (a) The distribution of the scores across different forecast weeks and regions (green triangle indicates mean). We employed the KS-test (SI~\ref{sec:KS-test_SI}) and observed that the VAR-dn model achieves better forecast performance relative to the ARIMA baseline.  (b) A comparison of the relative WIS score ($<1$ means a better models than the baseline ARIMA) of models for different regions of VA. (c) Coefficients of the VAR model corresponding to lag 1 and 2 of the VAL. (d) Phase-based summary of the WIS scores for individual models.}
\label{fig:wis_scores_sum}
\end{figure}

We compare rolling probabilistic forecasts from ARIMA, ARIMAX-Fct, VAR, and their denoised-VAL variants (ARIMAX-fct-dn, VAR-dn). Figure~\ref{fig:wis_scores_sum} summarizes performance during the post-PHE period. VAR-dn has the lowest WIS distribution and the most consistent relative WIS improvement across regions. Models using VAL-dn generally outperform their non-denoised counterparts, supporting the predictability analysis. The VAR coefficients associated with lagged VAL increase during periods of growth in hospital admissions, providing additional evidence that wastewater signals are most informative during surge dynamics.

Forecast coverage further supports the value of VAR-dn, particularly at longer horizons and during surge phases. For 4-week-ahead forecasts, VAR-dn achieves higher 95\% observed coverage than ARIMA and ARIMAX-Fct-dn across decline, plateau, and surge phases (Table~\ref{tab:phase_cov}). During surge periods, VAR-dn improves observed 95\% coverage by approximately 20 percentage points relative to the baselines. During plateau phases, the hospitalization dynamics are often relatively stable and strongly autoregressive, which can favor simpler univariate ARIMA model over the ARIMAX model. Also ARIMAX generally introduce additional uncertainty due to the estimation of cross-variable dependencies and the need to forecast future wastewater signals for multi-step prediction. Hence, we observe ARIMA to have competitive performance (both in terms of WIS and coverage) similar to the VAR-dn model during the plateau phase.

\begin{table}[h!]
\scriptsize
    \centering
    \begin{tabular}{llccc}
    \hline
        Model & Phase & Covered & Total & Observed coverage (\%) \\
        \hline
        ARIMA & Decline & 126 & 225 & 56.0 \\
        ARIMAX-Fct-dn & Decline & 97 & 225 & 43.1 \\
        VAR-dn & Decline & 164 & 225 & 72.9 \\
        ARIMA & Plateau & 18 & 28 & 64.3 \\
        ARIMAX-Fct-dn & Plateau & 9 & 28 & 32.1 \\
        VAR-dn & Plateau & 23 & 28 & 82.1 \\
        ARIMA & Surge & 72 & 167 & 43.1 \\
        ARIMAX-Fct-dn & Surge & 73 & 167 & 43.7 \\
        VAR-dn & Surge & 110 & 167 & 65.9 \\
        \hline
    \end{tabular}
    \caption{Observed 95\% coverage of 4-week-ahead forecasts by epidemic phase. VAR-dn provides higher coverage across all phases.}
    \label{tab:phase_cov}
\end{table}
\subsection{Effect of Wastewater Reporting Delays}
On several occasions we observed that the wastewater signal reporting would be delayed by 1$-$2 weeks due to multiple factors. The delay implies that, when forecasting at time $t$, we would not have access to $v(t)$. Here, we analyze the effects of delay in reporting on the forecast performance. We simulate two scenarios, a one-week and two-week delay in wastewater reporting. We modify the VAR model to account for the lack of availability of $v(t)$ and describe the model in SI~\ref{app:rep-delays}. VAR-dn-del1 (with $\delta=1$) and VAR-dn-del2 (with $\delta=2$) models simulate one- and two-week delay in wastewater reporting, respectively.

\begin{figure}[h!]
    \centering
       \subfloat[]{
        \includegraphics[width=0.45\textwidth]{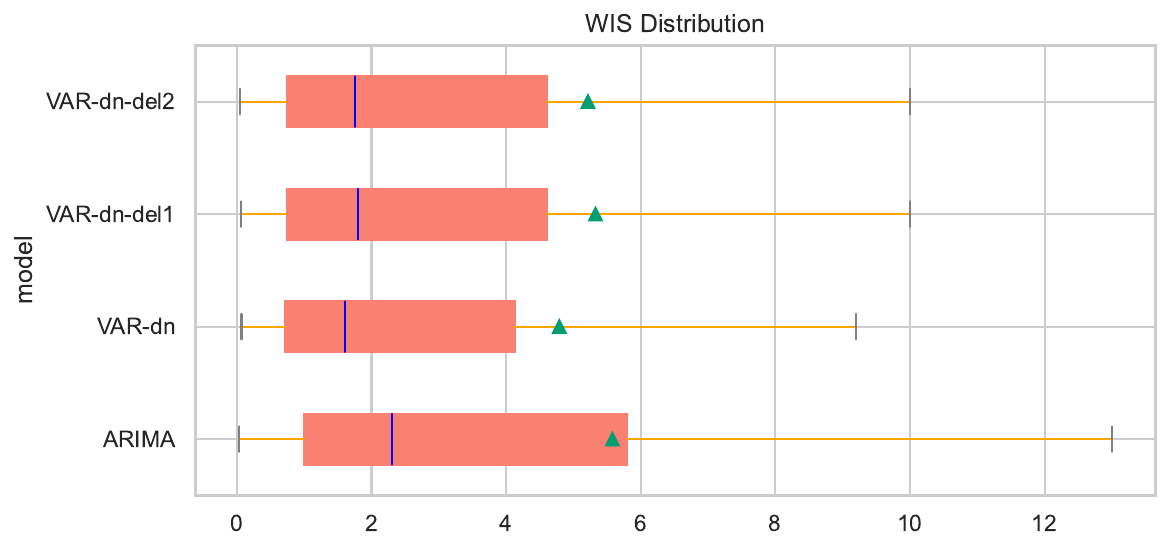}
        \label{fig:boxplot_wis_del}}
       \subfloat[]{
        \includegraphics[width=0.45\textwidth]{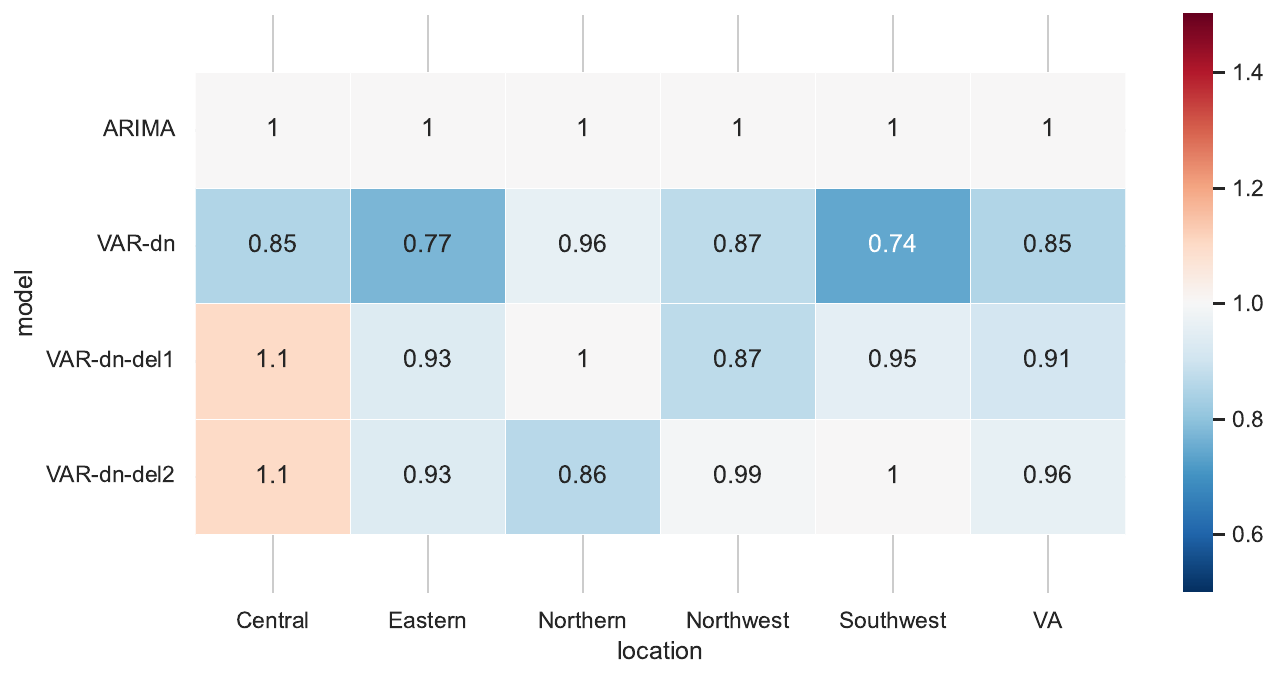}
        \label{fig:rel_heatmap_dn_del}}
        \\
\subfloat[]{\includegraphics[width=.55\textwidth]{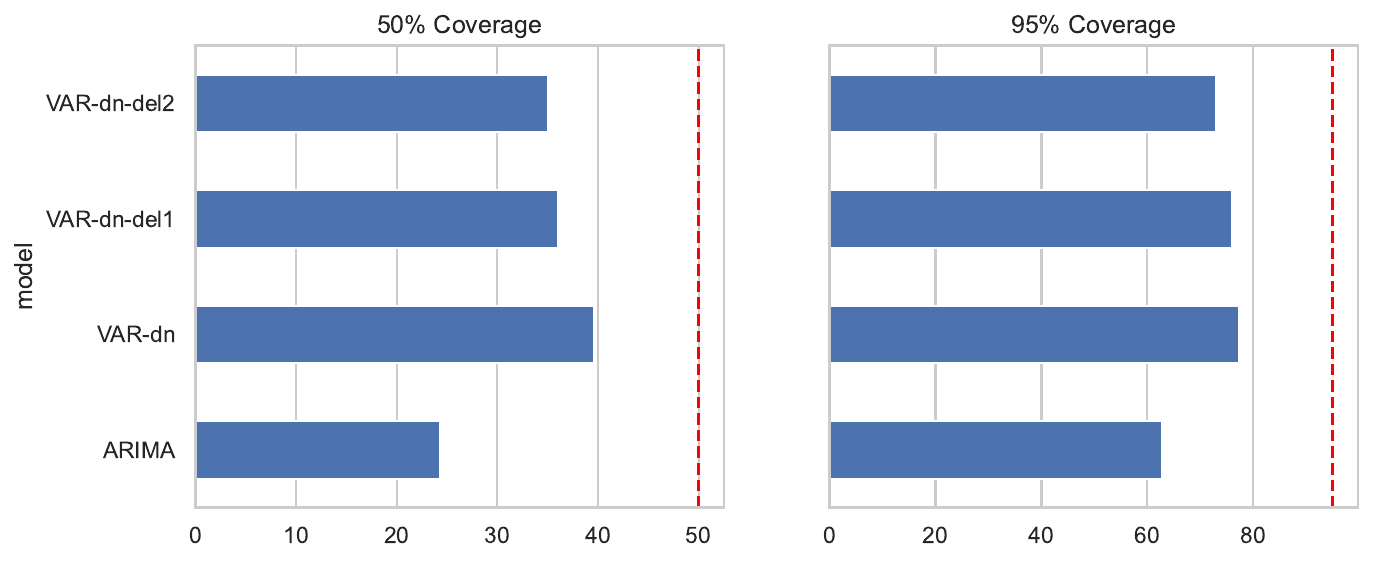}
        \label{coverage_del}}
    \caption{Simulating delayed wastewater reporting. We simulate two scenarios, a one-week and two-week delay in wastewater reporting. VAR-dn-del1 and VAR-dn-del2 models simulate one- and two-week delay in wastewater reporting, respectively.  (a) The distribution of the scores across different forecast weeks and regions (green triangle indicates mean). KS-test indicates (see Section~\ref{sec:KS-test_SI}) that the WIS distribution of the delayed VAR model was systematically smaller than that of the ARIMA reference model.  (b) Relative WIS scores for the various models. (c) 50\% and 95\% coverage of models. Across all three metrics, we observe that the VAR-dn-del1 and VAR-dn-del2 perform better than the ARIMA model. This indicates that despite the delay in reporting, the wastewater signal can still improve forecast performance.}
    \label{fig:del_wis_coverage}
\end{figure}

In Figure~\ref{fig:del_wis_coverage}, we summarize the WIS scores, relative WIS scores, and the coverage of the model forecasts. Comparing the performance of the VAR-dn-del1 and VAR-dn-del2 with the VAR model, we observe that employing the delayed VAL time series reduces the performance of the two models. However, we observe that the two models have a better performance than the ARIMA model (determined using the KS test to compare the WIS distribution of VAR-dn-del1 and VAR-dn-del2 with the WIS distribution of the ARIMA model). These results indicate that incorporating the delayed VAL time can still improve the forecast performance of the models. 

\section{Discussion and Conclusion}
\label{sec:conclusion_condensed}
We developed EpiFlow to evaluate and improve the forecasting utility of wastewater-based surveillance signals. Applied to COVID-19 hospital admissions in Virginia, the framework shows that wastewater-derived viral activity can improve probabilistic forecasts when the signals are appropriately processed, denoised, and modeled through time-varying dynamics. In our framework, the strongest gains are observed during surge periods, when wastewater signals often provide leading information about future hospital burden. 

The results highlight three methodological points. First, wastewater signals are noisy and their predictive structure depends strongly on the analysis window; shorter windows can retain more usable temporal structure than long histories. Second, denoising improves wastewater predictability and forecast performance. Third, the relationship between wastewater and hospital admissions is dynamic: rolling GC shows that wastewater can lead hospital admissions during some periods, while the relationship can weaken or reverse during others. This supports the use of rolling VAR models rather than static unidirectional models. Unlike latent-variable and state-space frameworks which model unobserved infection incidence while treating wastewater and hospitalization observations as noisy manifestations of the latent epidemic process, we explicitly separate denoising, rolling-window estimation, and forecasting to reduce model complexity and assumptions.

The framework has limitations. The VAR model assumes linear relationships and does not provide a mechanistic explanation for shedding, transport, or hospitalization processes. The analysis also assumes stable spatial aggregation of sewersheds and health regions, although real surveillance systems can experience site dropout and changing sampling coverage. Future work will incorporate mechanistic components, adaptive site weighting, and nonlinear or Bayesian state-space models to better capture uncertainty and evolving wastewater--burden relationships. Additionally, based on the causal relationship discussion (Conceptual causal graph, SI~\ref{sec:GC_all-SI}), we plan to explore the number of COVID-19 patients currently hospitalized (as opposed to new admissions) since it may exhibit stronger relationships with prevalence data and could provide better insights into wastewater dynamics.

\paragraph{EpiFlow Framework}
The data preprocessing, signal analysis, forecast modules presented in Section~\ref{sec:methods_condensed} are packaged as EpiFlow framework \url{https://github.com/aniruddhadiga/EpiFlow}. The details of the package are provided in SI~\ref{sec:epiflow_modules-SI}.

\paragraph{Data availability.}
The codes for EpiFlow and the data can be accessed at \url{https://github.com/aniruddhadiga/EpiFlow}.

\appendix



\section{Related Works}
\label{sec:related_works_SI}
First, we discuss recent work focused on improving wastewater-based disease surveillance and  wastewater-based disease forecasting. We also discuss how our work compares with these recent results, in particular as it pertains to forecasting results.

\paragraph{Data-quality enhancement methods} Dai et al.~\cite{dai2022statistical} have developed a Bayesian statistical framework based on functional principal component analysis to tackle the challenges of irregular sampling of sewershed and successfully detect the true trends of viral concentration out of noisy and sparsely observed viral concentrations. Leisman et al. ~\cite{leisman2024modeling} have developed a generalized pipeline using different correction models to reduce the variance in wastewater measurements but restrict the analysis to the Chicago area. Keshaviah et al.~\cite{keshaviah2022separating} developed an early warning algorithm called COVID-SURGE to reliably distinguish signal from noise in the wastewater data and identify the start of a surge. Manuel et al.~\cite{MANUEL2024174937} employed a Bayesian smoothing and forecasting model to process SARS-CoV-2 RNA concentrations in wastewater for short-term projection and surveillance. Watson et al.~\cite{watson2024jointly} employed a state-space model with sequential Monte Carlo methods to jointly estimate the effective reproduction number (R\textsubscript{eff}) and case ascertainment rate (CAR) using SARS-CoV-2 RNA concentrations in wastewater and reported case data. Miyazawa et al.~\cite{Miyazawa2024} employed a mechanistic SEIR model combined with an extended-Kalman-filter-based smoothing to estimate the effective reproduction number using SARS-CoV-2 RNA concentrations in wastewater, comparing them with notification-based estimates.   


\paragraph{Forecasting models} 
Rankin et al.~\cite{rankin2025multi} developed a generalized additive model to generate two types of categorical forecasts for hospitalization capacity risks and hospitalization rate trends using hospitalization and COVID-19 wastewater data for six US cities over the period January 2021 to November 2022. The model produces probabilistic forecasts 1$-$3 weeks ahead. The authors evaluate the performance at critical change points and show that including wastewater data positively impacts the model's performance. Karthikeyan et al. ~\cite{karthikeyan2021high}, employed an ARIMA model to produce point forecasts of new positive cases from the historical case data, wastewater data, and sample collection date in San Diego from July to October 2020 and show that Pearson correlation between the observed cases and predicted cases is high and also report a low Root Mean Squared Error. Cao and Francis~\cite{cao2021forecasting} employed a VAR model for point predictions of new cases using historical cases and viral load in Indiana (PA) from April 2020 to February 2021 and show that short-time series can reliably predict cases 1 week ahead. Zhao et al.~\cite{zhao2022five} compared the ARIMA model and VAR model using wastewater data from Detroit collected from September 2020 to August 2021 and showed that the VAR model is more effective in predicting COVID-19 incidence compared to the ARIMA model. In addition to the autoregressive models, other statistical models such as the GAM model, Poisson model, and Negative Binomial model have been investigated for predicting the COVID-19 cases from the wastewater data in the three New England regions in the US~\cite{anneser2022modeling}. Regression models have been considered for case prediction using wastewater data only. Joseph-Duran et al.~\cite{joseph2022assessing} similarly employed a simple regression model and   demonstrate its effectiveness for short-term case prediction. Aberi et al.~\cite{aberi2021quest} showed that different regression models, e.g. linear regression, polynomial regression, and generalized additive models can effectively forecast COVID-19 surveillance signals using the wastewater data using data from four treatment plants in Austria collected from May to December 2020. Klaassen et al.~\cite{KLAASSEN2024117395} employed a Bayesian nowcasting model and used multiple data sources, including SARS-CoV-2 RNA concentrations in wastewater, reported cases, deaths, and serosurvey data, to estimate infection trends and  R\textsubscript{t} across five sewersheds in Louisville, Kentucky. Their analysis demonstrated that wastewater data correlated with infection estimates and improved nowcasts when clinical surveillance data were limited, highlighting its potential for infectious disease monitoring in low-resource settings. Recent studies in wastewater-based epidemiology have increasingly emphasized dynamic and state-space modeling approaches for characterizing evolving relationships between wastewater viral concentrations and community infection dynamics. The CDC Center for Forecasting and Outbreak Analytics (CFA) developed a wastewater-informed forecasting framework that combines wastewater viral concentrations and hospital admissions data to infer latent infection dynamics and generate probabilistic forecasts of COVID-19 hospital admissions \cite{CDCWastewaterForecast2024}. Ouyang et al.~\cite{ouyang2025dynamic} employed dynamic linear state-space models to robustly estimate COVID-19 trends under missing and irregular wastewater observations, enabling probabilistic temporal updating and uncertainty quantification. Ensor et al.~\cite{ensor2025nonlinear} developed a nonlinear hierarchical time-series framework to model multiscale citywide viral trends and spatial hot spots across wastewater systems, while Sun et al.~\cite{sun2025uncovering} used Bayesian dynamic functional regression to capture time-varying and spatially heterogeneous relationships between wastewater viral concentrations and infection burden. Collectively, these studies highlight the utility of state-space, Bayesian, and dynamic time-series frameworks in modeling nonstationary epidemiological processes from wastewater surveillance data. Such approaches provide integrated alternatives to explicit preprocessing-based denoising incorporated in our framework while addressing the nonstationary and noisy nature of wastewater surveillance data.

 Several machine learning models, such as random forest, KNN, and deep neural networks, have also been considered in the literature. Li et al.~\cite{li2023wastewater} employed a random forest model and used multiple features in addition to viral loads, such as community vulnerability index~\cite{smittenaar2021covid}, vaccination coverage, population size, weather, and wastewater temperature to predict COVID-19-induced weekly new hospitalizations in 159 counties across 45 states in the US. Aberi et al.~\cite{aberi2021quest}, in addition to the regression models, also considered $k$-nearest neighbor, multilayer perceptron, support vector regression, decision trees, and random forest. Fu et al.~\cite{FU2024175830} employed a Gaussian model and random forest algorithm to predict the epidemic trajectories of SARS-CoV-2 and influenza A virus (IAV) using SARS-CoV-2 and IAV RNA concentrations in wastewater from three key port cities in China. Their model extracted key epidemic features from wastewater data and demonstrated the model's utility for real-time outbreak prediction and provided case trends up to six months ahead. Cañas Cañas et al.~\cite{2025FrESE} employed a Light Gradient Boosting Machine (LightGBM) model and used multiple features, including SARS-CoV-2 RNA concentrations in wastewater, clinical case counts, population size, and meteorological variables, to predict COVID-19 case trends across municipalities in Catalonia, Spain. Their model demonstrated strong generalization ability across different regions and epidemiological waves. Some deep learning-based models have also been explored for the wastewater-based epidemic surveillance tasks~\cite{zhu2022covid,jiang2022artificial,li2021data,galani2022sars,ZAMARRENO2024170367}. Specifically, the artificial neural network model (ANN) and adaptive neuro fuzzy inference system (ANFIS) have proven effective in different studies for case prediction tasks when compared with linear models and random forest~\cite{li2021data}.
 
 Recently, semi-mechanistic approaches~\cite{CDCWastewaterForecast2024, lisonAdrianEpiSewerPackage} have been developed for estimating effective reproductive number from WVL at sewershed and state level, which can then be used to forecast COVID-19 hospital admissions. Morvan et al.~\cite{morvan2022analysis} combined the shedding model and gradient-boosted regression trees (GBRT) to obtain highly accurate estimates (1.1\% of the estimates from the representative prevalence survey) of COVID prevalence in England with the wastewater data from 45 sewage sites. Meadows et al.~\cite{MEADOWS2025122671} employed a stochastic SEIR model with a particle filter method to predict COVID-19 outbreaks using SARS-CoV-2 RNA concentrations in wastewater from five rural communities and a small city in Idaho, USA. Their model successfully forecasted Omicron outbreaks with a lead time of 0 to 11 days before reported clinical cases, demonstrating the feasibility of wastewater-based surveillance for early outbreak detection in rural areas.

We provide a table summarizing the contributions, methods, forecast horizon, and study period in Table~\ref{tab:forecast_models-SI} of the papers for ease of comparison. Also, we refer the reader to the review article by Chen-Chen et al~\cite{chen2024wastewater} for a comprehensive review of the publications related to wastewater-based forecasting.
\begin{sidewaystable}
\centering
\small
\centering
\scriptsize
\begin{tabular}{|p{3cm}|p{3.2cm}|p{4cm}|p{3.5cm}|p{2.5cm}|}
\hline
\textbf{Authors} & \textbf{Type of Models} & \textbf{Type of Forecasts (with Horizon)} & \textbf{Study Period} & \textbf{Number of Locations Considered} \\
\hline
Rankin et al.~\cite{rankin2025multi} & Generalized Additive Model (Statistical) & Categorical, Probabilistic (1–3 weeks ahead) & Jan 2021 to Nov 2022 (PHE) & 6 US cities \\
\hline
Karthikeyan et al.~\cite{karthikeyan2021high} & ARIMA (Statistical) & Point (Not explicitly stated) & July – Oct 2020 (PHE) & 1 (San Diego) \\
\hline
Cao and Francis~\cite{cao2021forecasting} & VAR (Statistical) & Point (1 week ahead) & Apr 2020 – Feb 2021 (PHE) & 1 (Indiana, PA) \\
\hline
Zhao et al.~\cite{zhao2022five} & ARIMA, VAR (Statistical) & Point (short term) & Sep 2020 – Aug 2021 (PHE) & 1 (Detroit) \\
\hline
Anneser et al.~\cite{anneser2022modeling} & GAM, Poisson, Negative Binomial (Statistical) & Point (short term) & 4-7 months (2020, 2021) (PHE) & 3 (New England regions) \\
\hline
Joseph-Duran et al.~\cite{joseph2022assessing} & Regression (Statistical) & Point (Short-term) & Sep 2020–Mar 2021 & 32 (Catalonia, Spain) \\
\hline
Aberi et al.~\cite{aberi2021quest} & Linear, Polynomial, GAM (Statistical); KNN, MLP, SVR, DT, RF (Machine Learning) & Point (Short-term) & May – Dec 2020 & 4 (Austria) \\
\hline
Klaassen et al.~\cite{KLAASSEN2024117395} & Bayesian Nowcasting (Statistical) & Probabilistic (Nowcasting) & Aug 2020 - Mar 2021 (PHE) & 5 (Louisville, KY) \\
\hline
Li et al.~\cite{li2023wastewater} & Random Forest (Machine Learning) & Point (short term) & Jun 2021 - Jan 2023 (PHE) & 159 counties (45 US states) \\
\hline
Fu et al.~\cite{FU2024175830} & Gaussian (Statistical), Random Forest (Machine Learning) & Point (Up to 6 months ahead) & Feb 2023-Apr 2024 & 3 port cities (China) \\
\hline
Cañas Cañas et al.~\cite{2025FrESE} & LightGBM (Machine Learning) & Point (short term) & July 2020 - July 2022 & Not specified (Catalonia, Spain) \\
\hline
Zhu et al.~\cite{zhu2022covid}, Jiang et al.~\cite{jiang2022artificial}, Li et al.~\cite{li2021data}, Zamarreño et al.~\cite{ZAMARRENO2024170367} & ANN, ANFIS, Deep Learning (Machine Learning) & Point (Short-term; horizon varies by study) & Aug 2020 - Feb 2021, May 2020 - Dec 2021, Mar 2020 - May 2020, May 2021 - Jul 2023 & - \\
\hline
CDC~\cite{CDCWastewaterForecast2024} & Semi-mechanistic & Probabilistic &Feb 2024 - April 2024 & US state level\\
\hline
\end{tabular}
\caption{Summary of modeling approaches using wastewater data for COVID-19 forecasting.}
\label{tab:forecast_models-SI}
\end{sidewaystable}

\section{Virginia Wastewater Monitoring Facilities}
\label{sec:SI-WWsites}
\begin{figure}[ht!]
    \centering
\includegraphics[width=.75\textwidth]{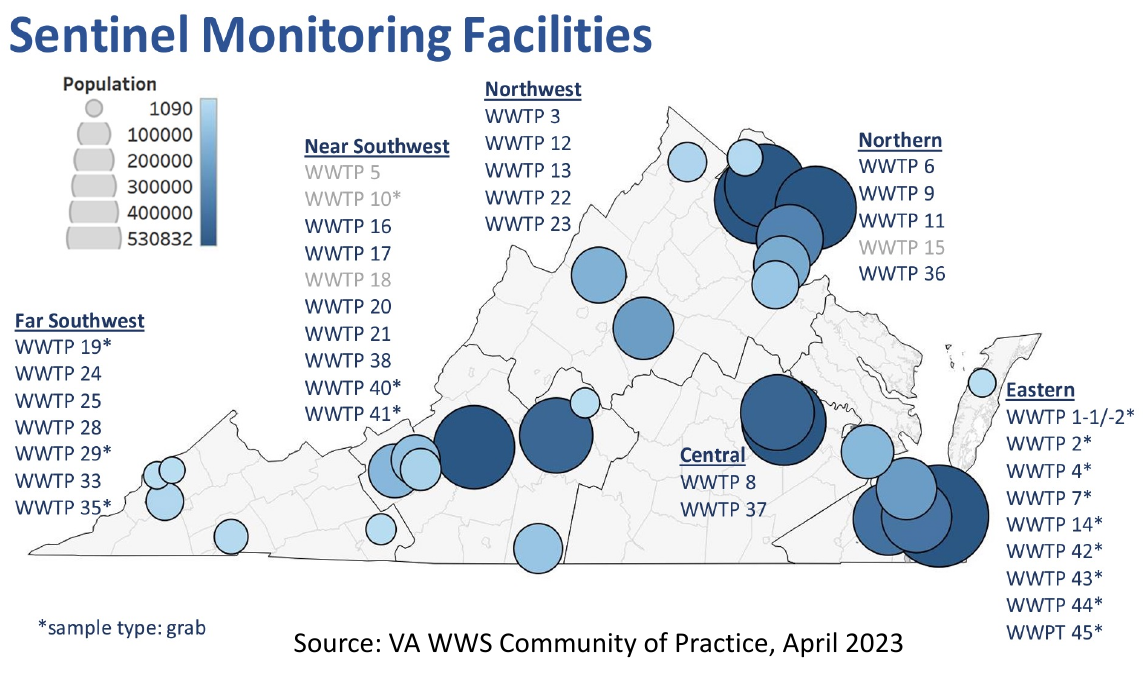}
    \caption{The geographic placement of the WWTPs across Virginia along with the population sizes served. In total, there are 36 WWTPs with 13 sites sampling twice weekly and 23 sites sampling once a week. Approximately, 50\% of the Virginia's population is monitored through these sites.}
    \label{fig:WWTP_VA}
\end{figure}

In Virginia, wastewater was collected each week from 36 wastewater treatment plants with 13 sites sampling twice weekly and 23 sites sampling once a week during our period of evaluation (October 2021$-$December 2023). The geographic distribution of the plants and population sizes served are shown in Figure~\ref{fig:WWTP_VA}. These samples help track the amount of SARS-CoV-2 in the wastewater. 
The concentration of SARS-CoV-2 is reported as “viral load” (WVL) $-$ the total amount of viral pieces entering the wastewater treatment plant on the day of sampling. It is calculated by analyzing the amount of SARS-CoV-2 virus pieces found along with the total daily flow at the treatment plant. Additional details about sampling techniques and viral load calculation are provided in~\cite{VDHdash,nwss}.
\section{Data Preprocessing Methods}
\label{sec:data_proc_methods_SI}
\subsection{Viral Activity Levels}
\label{sec:VAL-SI}
The VAL is a calculated measure that allows for the aggregation of wastewater sample data collected at the sewersheds to get regional, state, and national level trends. When multiple samples are collected in a week, the average of the logarithm of the WVL is computed prior to obtaining the VAL. The VAL is the number of standard deviations above the baseline, transformed to the linear scale. The baseline is the 10th percentile of the log-transformed and normalized concentration data within a specific time frame. The VAL at a specific geographic region for a particular week is computed by taking the median VAL of all the sewersheds serving the geographic region.

\subsection{Denoising Using Savitzky-Golay Filters}
\label{app:s-g_filters}
Savitzky-Golay filtering involves fitting a polynomial to local segments of the signal using a moving window approach, where the polynomial coefficients are determined via least squares regression. By appropriately selecting the window size and polynomial order, Savitzky-Golay denoising can strike a balance between noise reduction and signal fidelity. Unlike other denoising techniques, such as moving average or Gaussian smoothing, Savitzky-Golay denoising does not introduce significant phase distortion, making it particularly suitable for applications where preserving the temporal characteristics of the signal is crucial. This technique has been applied in various domains, including biomedical signal processing, spectroscopy, and chromatography, to enhance data quality and improve analysis outcomes. They have good spectral properties such as flat pass band and a steep cutoff frequency. SG filters also preserve signal features such as peaks better than any other smoothing filters. Also, the filter has a linear phase response and does not introduce any phase distortion. This property is particular important to ensure the lead-lag characteristics of the VAL signal is not altered.
\begin{figure}[h!]
    \centering
\includegraphics[width=.98\textwidth]{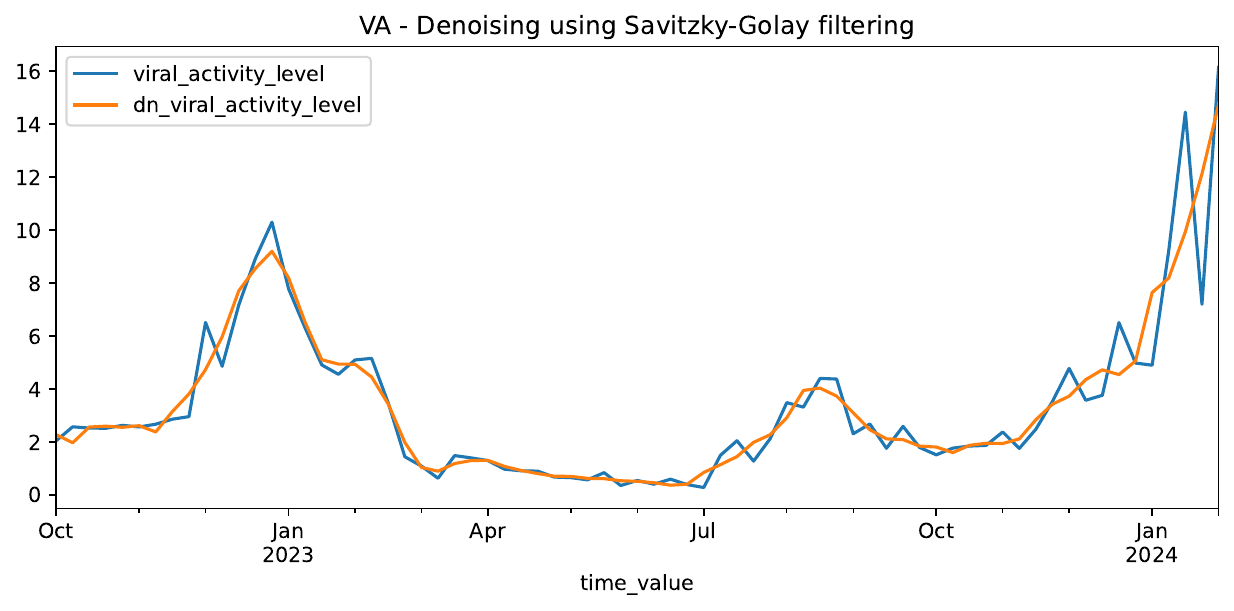}
    \caption{An example of denoising VAL timeseries corresponding to VA using Savitzky-Golay filters. Here the window size is 7 and the polynomial order is 2.}
    \label{fig:VA-denoising}
\end{figure}  

\section{Rolling-Window Granger Causality Test}
\label{sec:GC_all-SI}

\begin{figure}[ht!]
    \centering
        \centering
\subfloat[][]{\includegraphics[page=1,width=\textwidth]{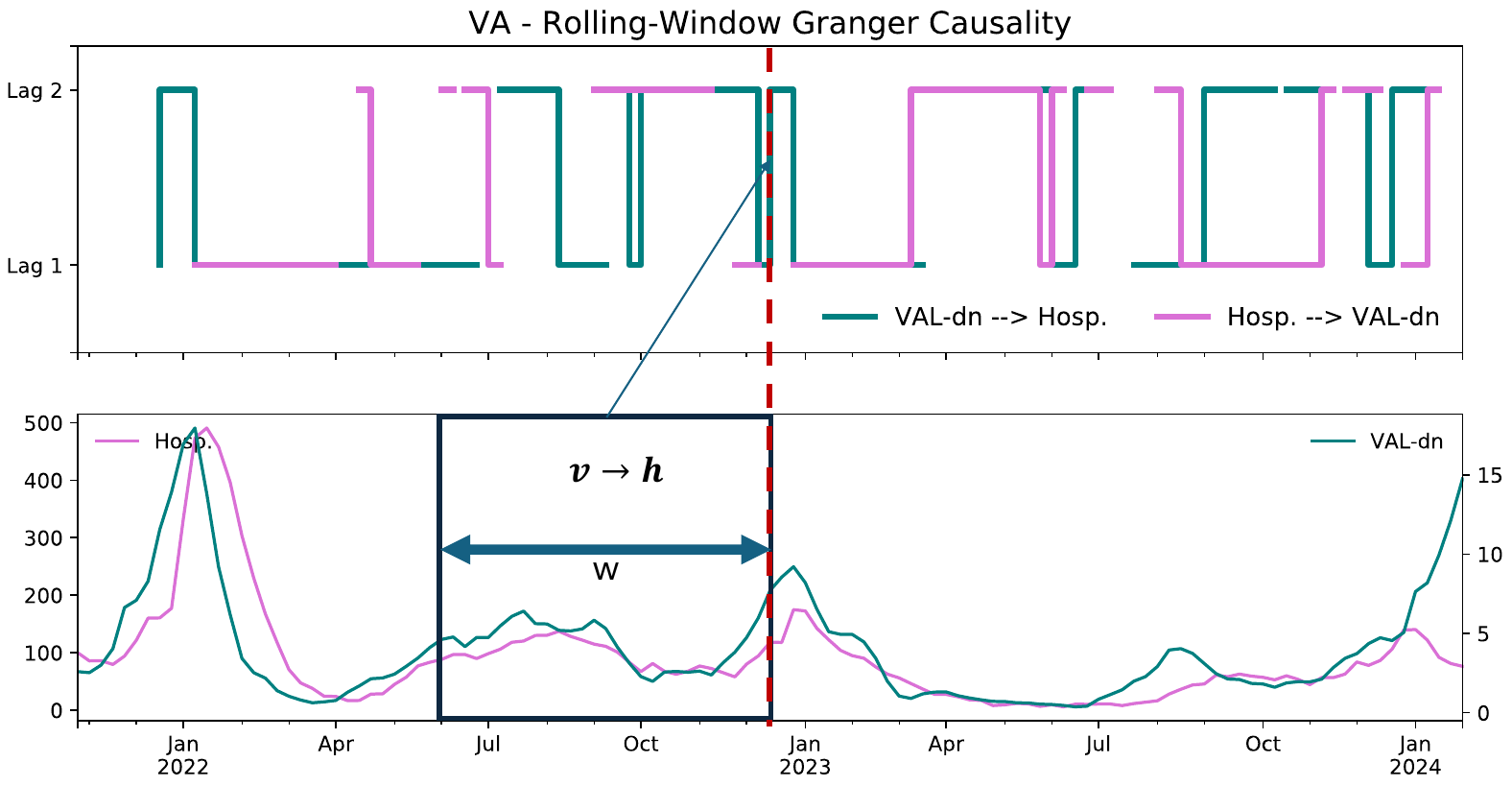}}
    \\
        \centering
\subfloat[][]{\includegraphics[page=2,width=\textwidth]{figs/rw-gc-test.pdf}}
    \caption{A description of the rolling-window GC. The red dashed line indicates the week for which the GC test is conducted and the black dashed box indicates the observation window over which the test is conducted. (a) shows a time-point where the $\mathbf{v}$ \emph{Granger causes} $\mathbf{h}$ and (b) shows a time-point where the $\mathbf{h}$ \emph{Granger causes} $\mathbf{v}$
    \label{fig:rwGC-example-SI}}
\end{figure}

\paragraph{Conceptual causal graph.} Figure~\ref{fig:wastewater_relationship} illustrates the conceptual relationship between infection prevalence, incident infections, wastewater concentrations, and hospital admissions. Here, prevalence refers to the total number of currently infected individuals within the population at a given time, whereas incidence denotes the number of new infections or hospital admissions occurring during a specified time interval. In this causal framework, wastewater RNA concentrations primarily reflect the prevalence of infected individuals due to sustained viral shedding, while new hospital admissions represent incident severe infections. Consequently, wastewater and hospitalization signals may exhibit indirect correlations/causations through their shared dependence on the latent prevalence process, while still providing useful predictive information for forecasting and early-warning analysis.
\begin{figure}[t]
\centering
\resizebox{\linewidth}{!}{
\begin{tikzpicture}[
    node distance=2.5cm and 3cm,
    every node/.style={font=\normalsize},
    state/.style={
        circle,
        draw,
        thick,
        minimum size=1.8cm,
        align=center
    },
    arrow/.style={
        -{Latex[length=3mm]},
        thick
    }
]

\node[state] (P) {$P(t)$};
\node[state, right=of P] (I) {$I[t,t+\Delta_1)$};
\node[state, below=of P] (W) {$W(t+\Delta_2)$};
\node[state, below=of I] (H) {$H[t,t+\Delta_3)$};

\draw[arrow] (P) -- (I);
\draw[arrow] (P) -- (W);
\draw[arrow] (I) -- (H);

\node[
    anchor=west,
    align=left,
    text width=7cm
] at ($(I.east)+(3.0,0)$)
{
    $P(t)$: Prevalence of infected individuals \\[0.6em]
    $I[t,t+\Delta_1)$: New infections (observed and unobserved) \\[0.6em]
    $W(t+\Delta_2)$: Wastewater concentrations \\[0.6em]
    $H[t,t+\Delta_3)$: New hospitalizations
};

\end{tikzpicture}
}
\caption{Conceptual relationship between prevalence, infections, wastewater concentrations, and hospital admissions with associated temporal delays.}
\label{fig:wastewater_relationship}
\end{figure}
\paragraph{An illustration of the Rolling-window GC test.} Figure~\ref{fig:rwGC-example-SI} provides an illustration of the rolling-window GC test applied on the VA state-level Hosp. and VAL. 
In Figure~\ref{fig:granger_causality_regions-SI}, we show the results of the rolling window GC test for the other five regions
    \begin{figure}[ht!]

        \centering
\subfloat[][]{       \includegraphics[width=0.9\textwidth]{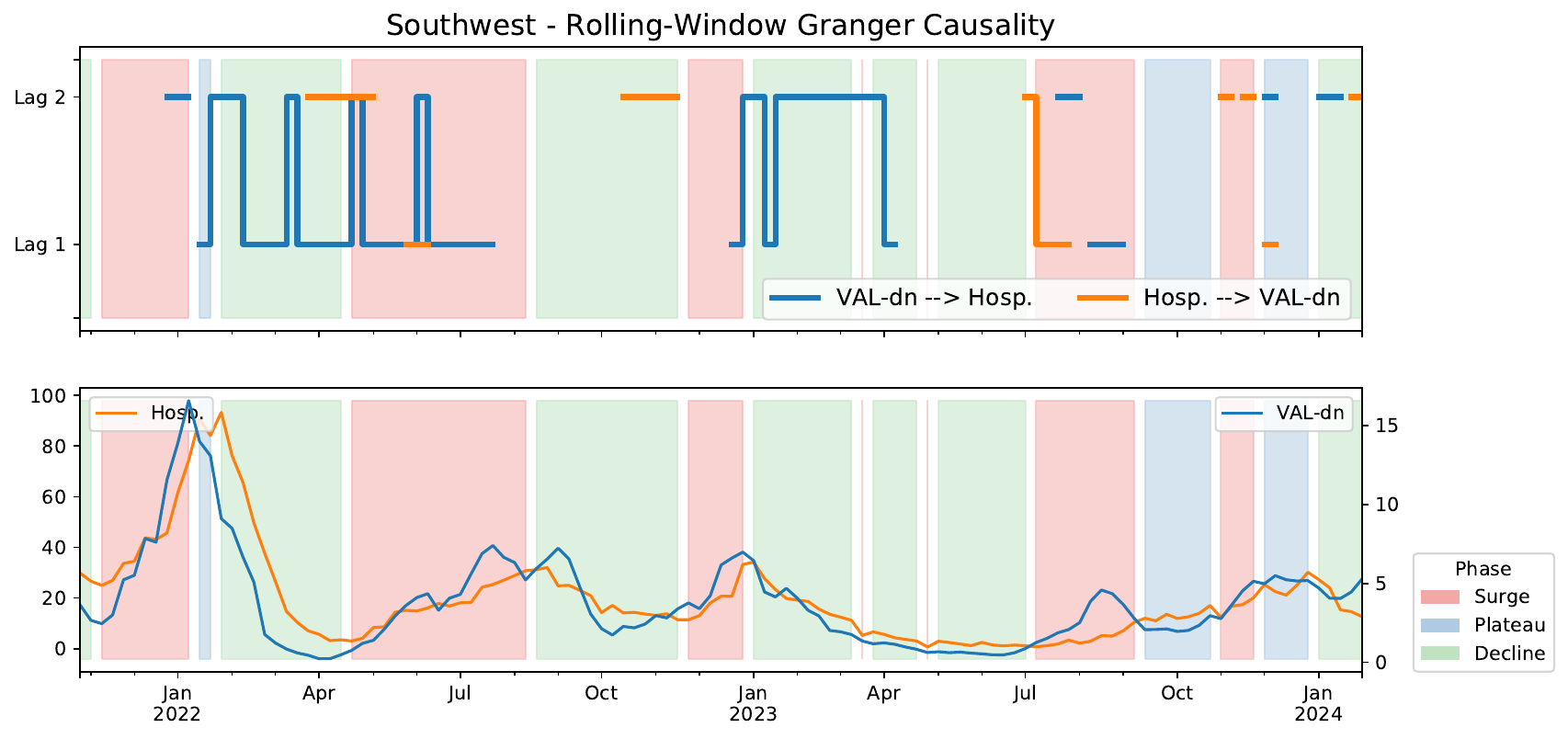}}
        \\
                \medskip

\subfloat[][]{        \includegraphics[width=0.9\textwidth]{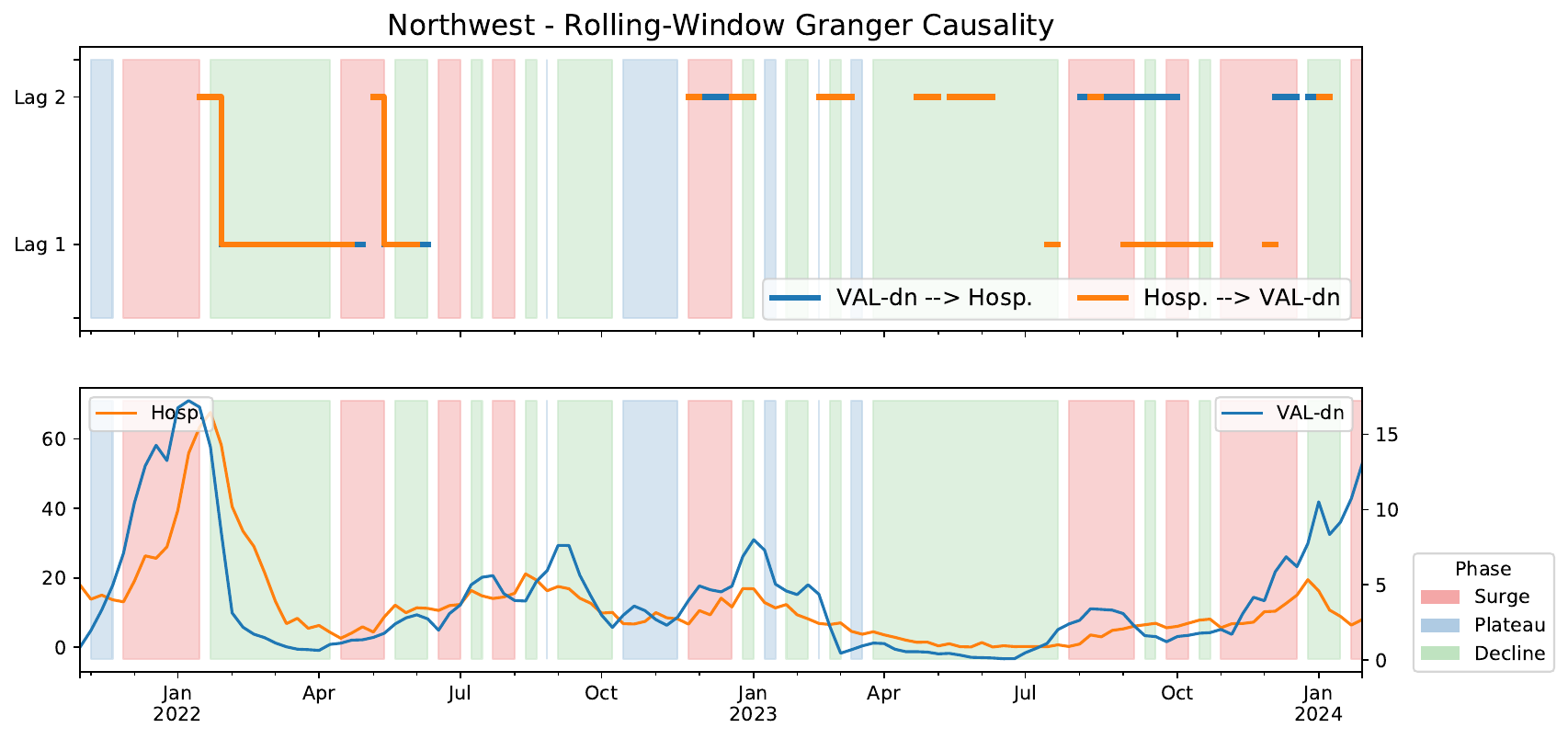}}
        \\
\subfloat[][]{        \includegraphics[width=0.9\textwidth]{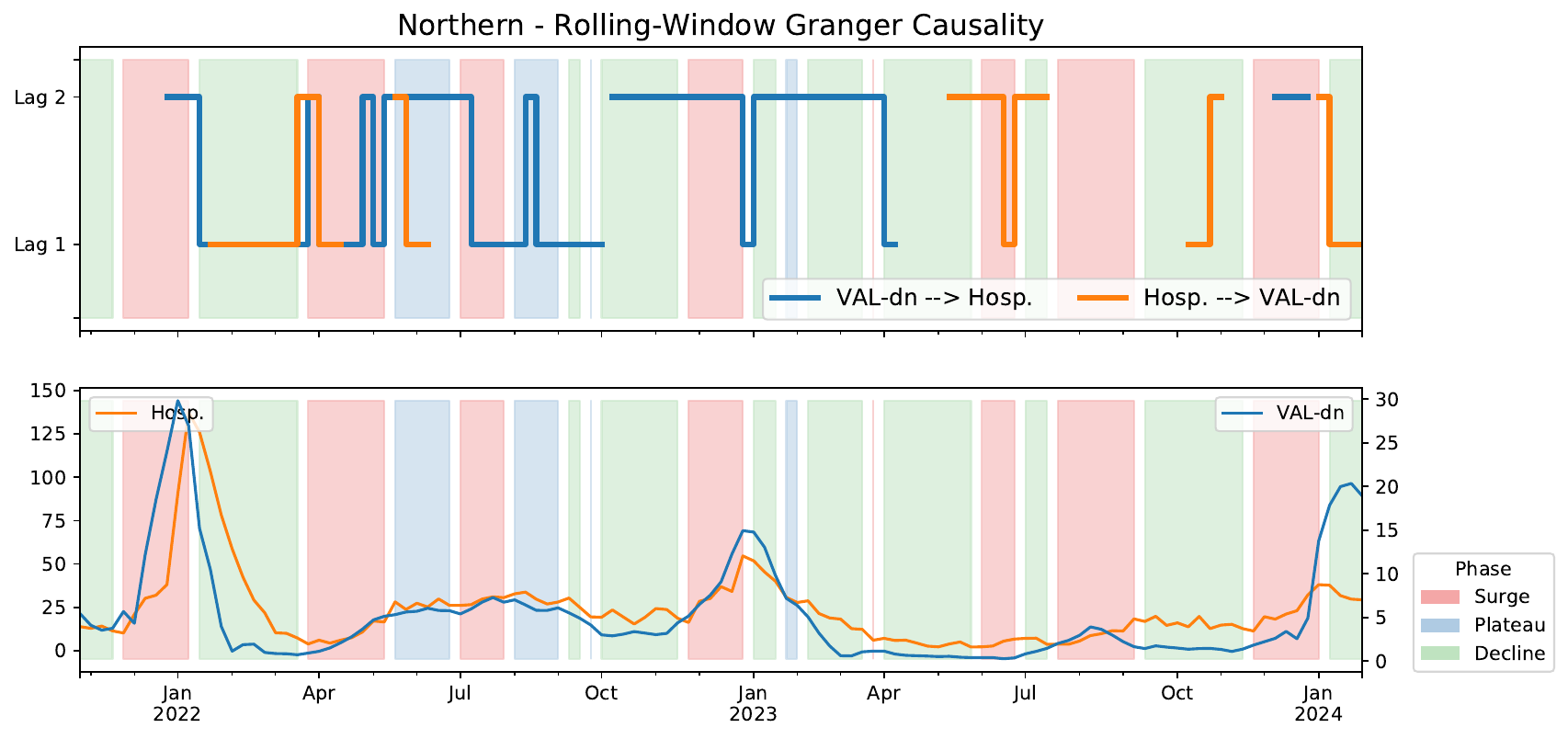}}
        \\
        \end{figure}%
\begin{figure}[ht!]
        \ContinuedFloat
\subfloat[][]{        \includegraphics[width=0.9\textwidth]{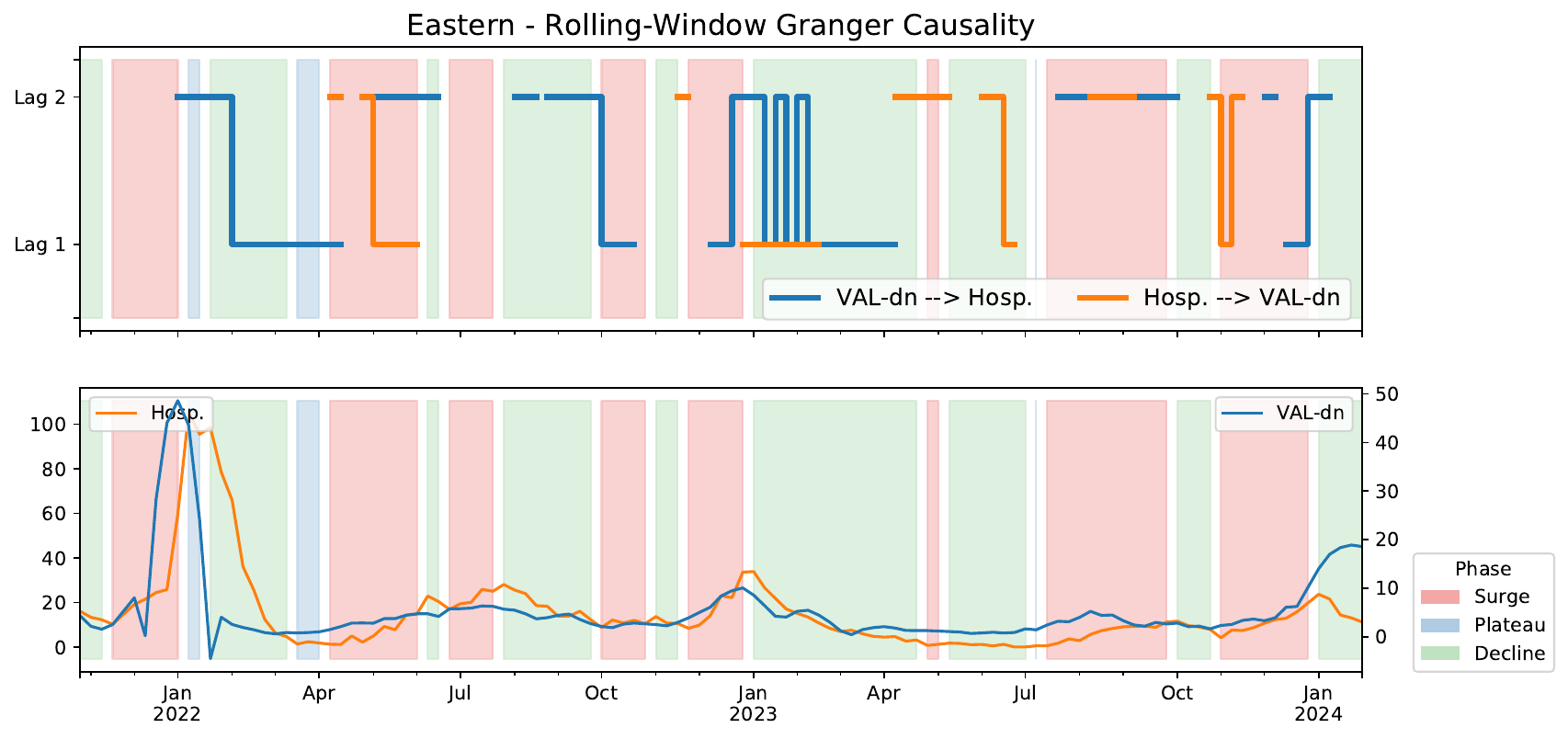}}
        \\
        \subfloat[][]{\includegraphics[width=0.9\textwidth]{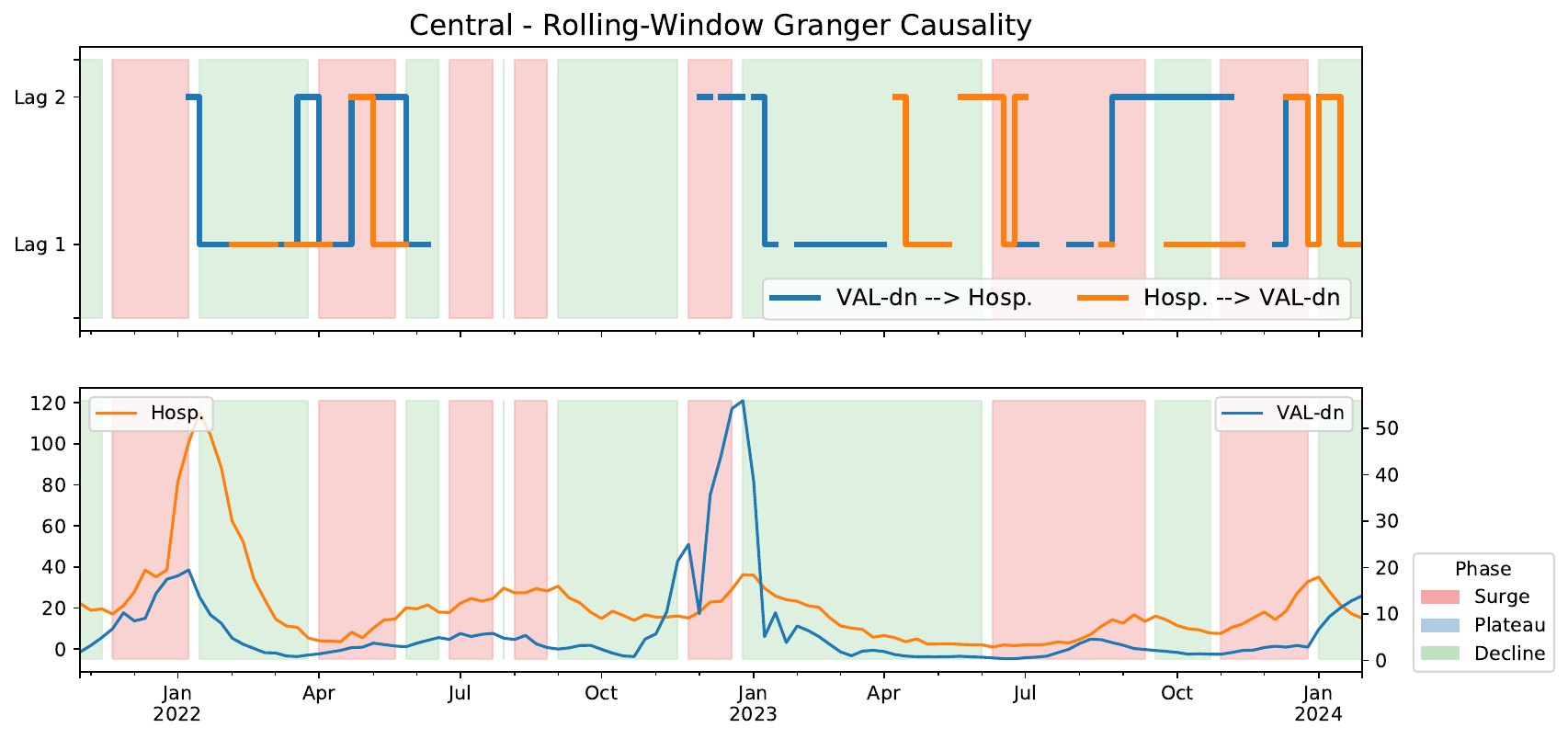}}
        \caption{The rolling-window Granger causality test performed on the VAL and Hosp. time series corresponding to the five VA health regions. In the top plot, the line color indicates the variable Granger causing the other variable and the value indicates the corresponding lag for that week. The bottom plot shows the VAL and Hosp. admission time series.}
        \label{fig:granger_causality_regions-SI}
    \end{figure}
\section{Additional Forecast Models}
\label{app:forecast_models}

\paragraph{Rolling-window ARIMA and ARIMAX forecasting framework.}
We additionally construct univariate autoregressive integrated moving average (ARIMA) and autoregressive integrated moving average with exogenous inputs (ARIMAX) models for forecasting hospitalization time series $h(t)$. While the ARIMA model utilizes only the historical dynamics of hospitalization counts, the ARIMAX model additionally incorporates wastewater-derived viral activity signals $v(t)$ as an external predictor.

The ARIMA($p,d,q$) model is given by
\begin{equation}
\phi(B)(1-B)^d h(t)
=
\theta(B)\epsilon(t),
\label{eq:arima}
\end{equation}
where $B$ denotes the backshift operator, $p$ is the autoregressive order, $d$ is the differencing order, $q$ is the moving-average order, $\phi(B)$ and $\theta(B)$ are autoregressive and moving-average polynomials, respectively, and $\epsilon(t)$ represents a zero-mean innovation process. The ARIMAX model extends this formulation by including lagged exogenous predictors:
\begin{equation}
\phi(B)(1-B)^d h(t)
=
\beta(B)v(t)
+
\theta(B)\epsilon(t),
\label{eq:arimax}
\end{equation}
where $\beta(B)$ denotes the transfer-function coefficients associated with the exogenous wastewater signal.

Both ARIMA and ARIMAX models are estimated within a rolling-window framework to account for nonstationary epidemic dynamics and evolving temporal relationships between hospitalization and wastewater indicators. For each rolling window, model parameters are re-estimated independently using only data available within the corresponding training interval. This enables the forecasting system to adapt to changing epidemic phases and transient relationships over time.

Model fitting is performed using the \texttt{statsmodels} implementation of the state-space ARIMA framework. Parameters are estimated through maximum likelihood estimation (MLE), where the likelihood of the observed time series under the assumed Gaussian innovation process is maximized numerically. Internally, the model is represented in state-space form and estimated using the Kalman filtering framework, enabling efficient recursive likelihood evaluation and forecasting.

Once fitted, the ARIMA and ARIMAX models generate probabilistic forecasts recursively across future horizons. The fitted state-space representation provides both predictive means and forecast error variances at each forecast horizon. Assuming approximately Gaussian forecast errors, the predictive distribution at horizon $h$ is represented as
\[
h(t+h)
\sim
\mathcal{N}
\left(
\hat{\mu}_{t+h},
\hat{\sigma}_{t+h}^2
\right),
\]
where $\hat{\mu}_{t+h}$ and $\hat{\sigma}_{t+h}^2$ denote the forecast mean and variance, respectively.

For each forecast horizon, probabilistic forecasts are constructed using the 23 quantile levels employed in the epidemic forecasting community.
For a quantile level $\tau$, the forecast quantile is computed as
$
\hat{q}_{\tau}
=
\hat{\mu}
+
z_{\tau}\hat{\sigma},$
where $z_{\tau}$ denotes the standard normal quantile corresponding to probability level $\tau$. The resulting set of predictive quantiles provides a probabilistic characterization of forecast uncertainty and variability across horizons.

The ARIMAX formulation enables direct incorporation of wastewater-derived viral activity indicators into the forecasting process, thereby allowing assessment of whether exogenous wastewater signals improve predictive performance relative to hospitalization-only autoregressive models.

\section{Predictability and Rolling Window Analysis}
\label{sec:predictability-SI}
Here we provide additional results of the predictability analysis discussed in Section~\ref{sec:predictability_results}. The goal of this analysis is to determine the cutoff on the analysis window length of VAL, beyond which the increasing window length does not change the mean predictability significantly. Given a region's time series, we compute the distribution of predictability for each window length. We employ Tukey's Honest Significant Difference (HSD) test to compare the means of all pairs of distributions. Given two means $\mu_i$ and $\mu_j$ of the distributions corresponding to window lengths $i$ and $j$, respectively, Tukey's HSD tests considers the null hypothesis $H_0: \mu_i=\mu_j$ and the alternate hypothesis $H_\alpha: \mu_i \neq \mu_j$. The significance level $\alpha$ was set to 5\%. 
\paragraph{Procedure.} 
\begin{itemize}
    \item Compute the pairwise Tukey's HSD between the predictability distribution corresponding to two different window lengths $i$ and $j$. Let $HSD(i,j)=\begin{cases}
    1, \quad \text{if $H_0$ rejected}\\
    0, \quad \text{else.}
\end{cases}$.
\item Determine the smallest $i$ such that, $HSD(i,j)=0, \, \forall j \ge i$. 
\end{itemize}
This $i$ serves as the cutoff observation window length in all our analysis.

\begin{figure}
    \centering
    \includegraphics[width=0.95\linewidth]{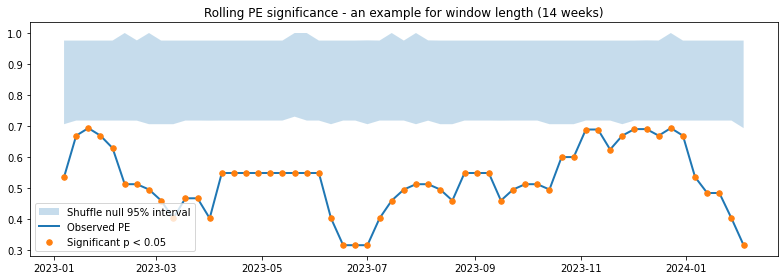}
\caption{Rolling-window permutation entropy (PE) significance analysis for an example window length of 14 weeks. The blue curve denotes the observed normalized permutation entropy computed within each rolling window, while the shaded region represents the 95\% interval of the surrogate entropy distribution obtained through random shuffling of the time series. Orange markers indicate windows where the observed entropy is significantly lower than the shuffle-based surrogate distribution ($p<0.05$), suggesting the presence of statistically significant temporal structure and increased predictability.}    \label{fig:perm_sig_test-SI}
\end{figure}

\paragraph{Permutation-based significance testing for permutation entropy} Figure~\ref{fig:perm_sig_test-SI} illustrates the rolling-window permutation entropy (PE) significance analysis for an example window length of 14 weeks. For each rolling window, the observed normalized PE (blue line) is compared against an empirical null distribution generated through random shuffling of the time series, which preserves the marginal distribution while destroying temporal dependencies. The shaded region represents the 95\% interval of the surrogate (shuffle-based) entropy distribution, while orange markers indicate windows where the observed PE is significantly lower than the surrogate distribution ($p<0.05$), suggesting the presence of statistically significant temporal structure and increased predictability within the corresponding time interval.

\begin{figure}[h!]
    \centering
\includegraphics[width=.75\textwidth]{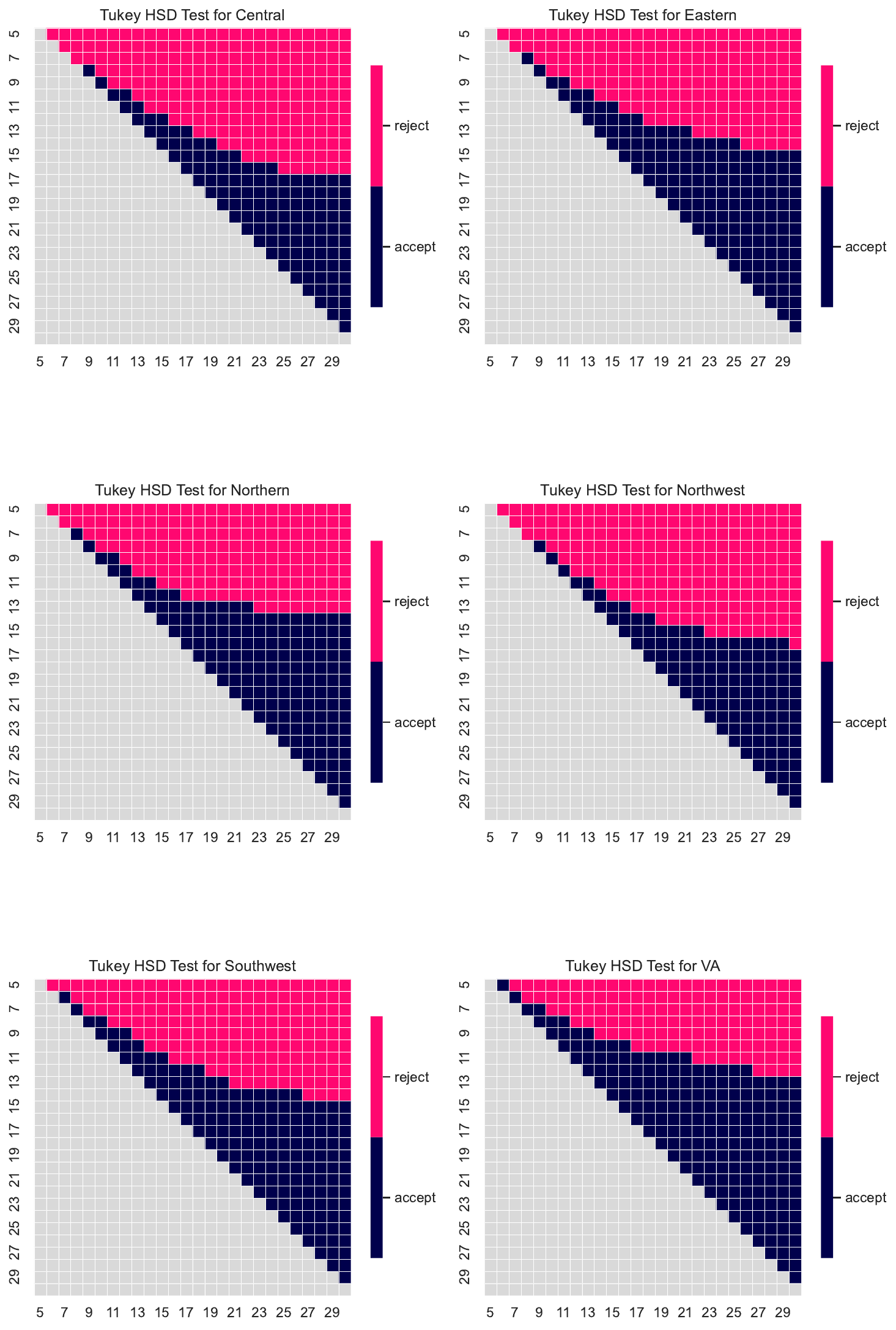}
    \caption{Understanding the difference in mean predictability across different rolling-window lengths for different locations using Tukey Honest Significant Difference Test ($\alpha=0.05$)}
    \label{fig:pred_rolling_tukeyHSD}
\end{figure}

\paragraph{Tukey-HSD  examples} The $HSD$ matrix computed for all the six locations is shown in Figure~\ref{fig:pred_rolling_tukeyHSD}. In these plots, we only show results for window lengths ranging from 5$-$30 weeks. Beyond 30, we did not observe significant changes in predictability.

\section{Kolmogorov-Smirnov test}
\label{sec:KS-test_SI}
The two-sample Kolmogorov--Smirnov (KS) test is a nonparametric statistical test used to determine whether two samples originate from the same underlying distribution \cite{ Massey1951}. Let $F(x)$ and $G(x)$ denote the empirical cumulative distribution functions (ECDFs) of two samples. The KS statistic is defined as
\[
D
=
\sup_x
|F(x)-G(x)|,
\]
where $\sup_x$ denotes the maximum absolute difference between the two ECDFs over all possible values of $x$. Here, $x$ represents the observed values of the quantity being compared (e.g., weighted interval score (WIS), permutation entropy, or forecast error values).

In the one-sided KS test with alternative hypothesis \texttt{alternative='less'}, the null and alternative hypotheses are given by
\[
H_0: F(x)=G(x),
\]
\[
H_1: F(x) < G(x),
\]
where the inequality is interpreted pointwise over the cumulative distributions. In practice, this tests whether the sample associated with $F(x)$ tends to produce systematically smaller values than the sample associated with $G(x)$. When applied to WIS distributions, a statistically significant result under the ``less'' alternative indicates that the empirical model produces lower WIS values and therefore better probabilistic forecasting performance relative to the reference model. We used the Python function \texttt{scipy.stats.kstest} function with \texttt{alternative=`less'} (\url{https://docs.scipy.org/doc/scipy/reference/generated/scipy.stats.kstest.html}) to test the hypothesis.

\section{Probabilistic Forecast Evaluation Metrics}
\label{app:forecast_eval}
\paragraph{Weighted Interval Score. }
In order to compare the forecast quantiles of the different models, we use the Weighted Interval Score (WIS), the de facto standard in epidemiological forecasting community for probabilistic forecast evaluation \cite{bracher2020evaluating}:
\begin{align}
    &WIS_{\alpha_{0:k}} (F,y) = \nonumber\\&\frac{1}{K+0.5} \sum_{k=0}^{K} \frac{\alpha_k}{2} (u_k-l_k) + \frac{2}{\alpha_k} (l_k-y) \mathbbm{1}(y<l_k) + \frac{2}{\alpha_k} (y-u_k) \mathbbm{1}(y>u_k) \nonumber\\
     &\qquad\qquad\qquad\text{dispersion}\quad+\quad\text{under prediction}\quad+\quad\text{over prediction}
    \label{eq:wis}
\end{align}
where $y$ is the observed value (ground truth case count corresponding to a week) for a given location and date, $F$ is the forecast defined in terms of the median $m$, upper quantiles $u_k$ and lower quantiles $l_k$ of the predictive distribution, respectively. $K$ is the number of intervals considered, which in our case $K=11$.

\paragraph{Relative WIS.} It is another metric used for evaluating forecast performance of COVID-19 forecast hub submissions~\cite{cramer2022evaluation,lopez2023predictive, sherratt2023predictive}. It is used to rank models based on pairwise comparisons between all models. The relative skill $\theta_i$
 of model $i$ is the geometric mean of all mean score ratios that involve model $i$. It is computed as $\theta_{i}=\left(\prod_{j=1}^{M} \theta_{ij}\right)^{1/M}$, where $\theta_{ij}=\frac{\text{average WIS of model } i}{\text{average WIS of model } j}$. Further, $\theta_i$ is scaled by the $\theta_B$, the relative skill of a baseline model $B$, to obtain the rescaled relative skill $\theta_i^*=\frac{\theta_i}{\theta_B}$. In our evaluation, ARIMA was considered as the baseline model.

 \paragraph{Coverage.}
 It is an estimate of the probability that a forecast interval (at a certain nominal level such as 80\%) correctly includes the actual value. It is estimated on a particular date by computing the proportion of locations for which a forecaster’s interval includes the actual value on that date. A perfectly calibrated forecaster would have each interval’s empirical coverage matching its nominal coverage. If the estimated coverage is lower than the nominal coverage, then the model is considered to be overconfident. In the case the estimated coverage is greater than the nominal coverage, the model is considered underconfident.

    \section{Phase Classification Algorithm}
    \label{sec:phase-class}
Let the given time series be $\mathbf{y}=[y(1), y(2), \cdots, y(t)]$. The algorithm to classify each time point in $\mathbf{y}$ is as follows:
\begin{enumerate}
    \item Approximate $\mathbf{y}$ with a piece-wise linear function to obtain breakpoints $\{b_1, \dots, b_m\}$ (the endpoints of a line segment) and the slope of the line segments.
    \item Phase classification rules:
    \begin{itemize}
        \item Time interval $(b_k, b_{k+1}]$ as \emph{surge} phase if the value at the second breakpoint $y_{b_{k+1}}$ exceeds $(1+\delta)$ times the value at the first breakpoint $y_{b_k}$, indicating an increasing trend in the time series from $b_k$ to $b_{k+1}$.
        \item Time interval $(b_k, b_{k+1}]$ as \emph{decline} phase if the value at the second breakpoint $y_{b_{k+1}}$ is less than $(1-\delta)$ times the value at the first breakpoint $y_{b_k}$, indicating a  decreasing trend in the time series from $b_k$ to $b_{k+1}$.
        \item If neither of the conditions holds, the time interval $(b_k, b_{k+1}]$ is categorized as a \emph{Plateau} phase.
    \end{itemize}
\end{enumerate}
For the piecewise linear approximation, we use a standard R package \verb|segmented| \cite{Rsegmented}. 

\section{Effects of Wastewater Reporting Delays: Forecast Models}
\label{app:rep-delays}
Here, we provide a description of the modified VAR and ARIMAX model accounting for the delays in wastewater reporting. Assuming a reporting delay of $\delta>0$ weeks, for the forecasting week $t$, v(t) is unavailable and only samples $[v(t-\delta), v(t-\delta-1),\cdots]$ is available, we modify \eqref{eq:var_granger} and rewrite it as
\begin{align}
     \begin{bmatrix}
         h(t)\\v(t-\delta)
     \end{bmatrix} = \begin{bmatrix}
         c_h\\c_v
     \end{bmatrix} + \begin{bmatrix}
         a^{(1)}_{hv} 
 a^{(1)}_{12}\\
         a^{(1)}_{vh} 
 a^{(1)}_{22}
     \end{bmatrix}
     \begin{bmatrix}
         h(t-1)\\v(t-\delta-1)
     \end{bmatrix}
     + \begin{bmatrix}
         a^{(2)}_{hh} 
 a^{(2}_{hv}\\
         a^{(2)}_{vh} 
 a^{(2)}_{vv}
     \end{bmatrix} 
     \begin{bmatrix}
         h(t-2)\\v(t-\delta-2)
     \end{bmatrix} + \ldots \nonumber\\+ \begin{bmatrix}
         a^{(p)}_{hh} 
 a^{(p)}_{hv}\\
         a^{(p)}_{vh} 
 a^{(p)}_{vv}
     \end{bmatrix} 
     \begin{bmatrix}
         h(t-p)\\v(t-\delta-p)
     \end{bmatrix} + \textbf{e}(t)
     \label{eq:var_eqn_del}
\end{align}

\paragraph{ARIMAX-Fct.} We also modify the ARIMAX model in \eqref{eq:arimax} as follows:
\begin{align}
    \quad (1 - \phi_1 L - \phi_2 L^2 - \ldots - \phi_p L^p) (1 - L)^d h(t) = \beta_0 + \beta_1 v(t-\delta) + \beta_2 v(t-\delta-1) + \ldots \nonumber\\+ \beta_k v(t-p-\delta)+ (1 + \theta_1 L + \theta_2 L^2 + \ldots + \theta_q L^q) \epsilon_t
\end{align}

\section{EpiFlow Software Package}
\label{sec:epiflow_modules-SI}
EpiFlow codebase is provided in~\url{https://github.com/aniruddhadiga/EpiFlow}. It contains the following modules:

\paragraph{Data Preprocessing Module} - This module primarily focuses on formatting time series data and ensuring temporal and spatial alignment across multiple datasets, facilitating accurate comparison. Key steps include data cleansing and harmonization of temporal and spatial dimensions. The necessary functions are provided in \texttt{viral\_utils.py}. An example of the of the use of the functions in data preprocessing workflow is provided in
\texttt{data\_preprocessing\_workflow.ipynb}
\paragraph{Signal Analysis Module} - This module focuses on ($i$) denoising, ($ii$) determining the appropriate window length based on predictability, and ($iii$) the causal relationship between the time series. The necessary functions for denoising are provided in \texttt{data\_proc\_utils.py}. Permutation entropy and Granger causality analysis functions are provided in \texttt{ww\_analyzer.py}. An example of the use of the functions in the signal analysis workflow is provided in
\texttt{signal\_analysis\_workflow.ipynb}
\paragraph{Forecasting module} - This module generates forecasts using a VAR model and is designed to provide real-time forecasts. The rolling-window length and the number of lags are determined from the signal analysis module, and the model is retrained when new observation is obtained. The model-related functions are provided in \texttt{VAR.py}. An example of the of the use of the functions in forecasting workflow is provided in
\texttt{forecast\_workflow.ipynb}.




\bibliographystyle{RS}
\bibliography{refs}

\end{document}